\documentclass{article}

\usepackage[nonatbib, final]{neurips_2022}
\usepackage[numbers]{natbib}
\usepackage[utf8]{inputenc} 
\usepackage[T1]{fontenc}    
\usepackage{float}
\usepackage{graphicx}
\usepackage[dvipsnames,table]{xcolor}
\usepackage{subcaption}
\usepackage{booktabs} 
\usepackage{tabularx}
\newcolumntype{Y}{>{\centering\arraybackslash}X}
\newcolumntype{Z}{>{\raggedleft\arraybackslash}X}
\usepackage{bm}
\usepackage{amsmath,amssymb,mathtools}      
\usepackage{nicefrac}
\usepackage[inline, shortlabels]{enumitem}
\usepackage{hyperref}
\usepackage{url}
\usepackage[capitalize,noabbrev]{cleveref}
\usepackage{makecell}
\usepackage{multirow}

\newcommand{\xhdr}[1]{\textbf{#1.}\;}

\newcommand{\bR}{\ensuremath \mathbb{R}}
\newcommand{\bN}{\ensuremath \mathbb{N}}

\newcommand{\nbinops}{\ensuremath N_{\mathrm{bin}}}
\newcommand{\nunops}{\ensuremath N_{\mathrm{una}}}
\newcommand{\psymb}{\ensuremath p_{\mathrm{sym}}}
\newcommand{\pconst}{\ensuremath p_{\mathrm{c}}}
\newcommand{\nconsts}{\ensuremath N_{\mathrm{const}}}
\newcommand{\nivs}{\ensuremath N_{\mathrm{iv}}}
\newcommand{\ngrid}{\ensuremath N_{\mathrm{grid}}}
\newcommand{\maxtime}{\ensuremath T}
\newcommand{\ivrange}{\ensuremath (y_0^{\min}, y_0^{\max})}

\newcommand{\neval}{\ensuremath N_{\mathrm{eval}}}
\newcommand{\nnodes}{\ensuremath K}
\newcommand{\pint}{\ensuremath p_{\mathrm{int}}}
\newcommand{\preal}{\ensuremath p_{\mathrm{real}}}

\newcommand*\samethanks[1][\value{footnote}]{\footnotemark[#1]}

\title{Discovering ordinary differential equations that govern time-series}

\author{%
  Sören Becker \thanks{equal contribution}\\
  Helmholtz AI, Munich \\
  \And
  Michal Klein \samethanks\phantom{n}\thanks{work was done while at Technical University of Munich.}\\
  Apple \\
  \And
  Alexander Neitz \thanks{work was done while at Max Planck Institute for Intelligent Systems.}\\
  DeepMind \\
  \And
  Giambattista Parascandolo \samethanks\\
  OpenAI \\
  \And
  Niki Kilbertus \\
  Helmholtz AI, Munich \& \\
  Technical University of Munich \\
}

\graphicspath{{./figs/}}

\begin{document}

\maketitle

\begin{abstract}
  Natural laws are often described through differential equations yet finding a differential equation that describes the governing law underlying observed data is a challenging and still mostly manual task. In this paper we make a step towards the automation of this process: we propose a transformer-based sequence-to-sequence model that recovers scalar autonomous ordinary differential equations (ODEs) in symbolic form from time-series data of a single observed solution of the ODE. Our method is efficiently scalable: after one-time pretraining on a large set of ODEs, we can infer the governing laws of a new observed solution in a few forward passes of the model. Then we show that our model performs better or on par with existing methods in various test cases in terms of accurate symbolic recovery of the ODE, especially for more complex expressions.
\end{abstract}

\section{Introduction}
\label{sec:introduction}

\emph{Science} is commonly described as the ``discovery of natural laws through experimentation and observation''.
Researchers in the natural sciences increasingly turn to machine learning (ML) to aid the discovery of natural laws from observational data alone, which is often abundantly available, hoping to bypass expensive and cumbersome targeted experimentation.
While there may be fundamental limitations to what can be extracted from observations alone,
recent successes of ML in the entire range of natural sciences provide ample reason for excitement.
In this work, we focus on ordinary differential equations, a ubiquitous description of dynamical natural laws in physics, chemistry, and systems biology.
For a first order ODE $\dot{y} := \nicefrac{\partial y}{\partial t} = f(y, t)$, we call~$f$ (which uniquely defines the ODE) the underlying dynamical law.
Informally, our goal is then to infer~$f$ in symbolic form given discrete time-series observations of a single solution $\{y_i := y(t_i)\}_{i=1}^n$ of the underlying ODE.

Contrary to ``black-box-techniques'' such as Neural Ordinary Differential Equations (NODE)~\citep{chen2018neural} that aim at inferring a possible~$f$ as an arguably opaque neural network, we focus specifically on symbolic regression.
From the perspective of the sciences, a law of nature is useful insofar as it is more broadly applicable than to merely describe a single observation.
In particular, the reason to learn a dynamical law in the first place is to dissect and understand it as well as to make predictions about situations that differ from the observed one.
From this perspective, a symbolic representation of the law (in our case the function~$f$) has several advantages over block-box representations: they are compact and directly interpretable, they are amenable to analytic analysis, they allow for meaningful changes and thus enable assessment of interventions and counterfactuals.

In this work we present NSODE, a sequence-to-sequence transformer that maps observed trajectories, i.e., numeric sequences of the form $\{(t_i, y_i)\}_{i=1}^n$, directly to symbolic equations as strings, e.g., \texttt{"y**2+1.64*cos(y)"}, which is the prediction for $f$.
This example directly highlights the benefit of symbolic representations in that the $y^2$ and $\cos(y)$ terms tell us something about the fundamental dynamics of the observed system; the constant \texttt{1.64} will have semantic meaning in a given context and we can, for example, make predictions about settings in which this constant takes a different value.

\begin{figure*}
  \centering
  \vspace{-7mm}
  \includegraphics[width=1.0\textwidth]{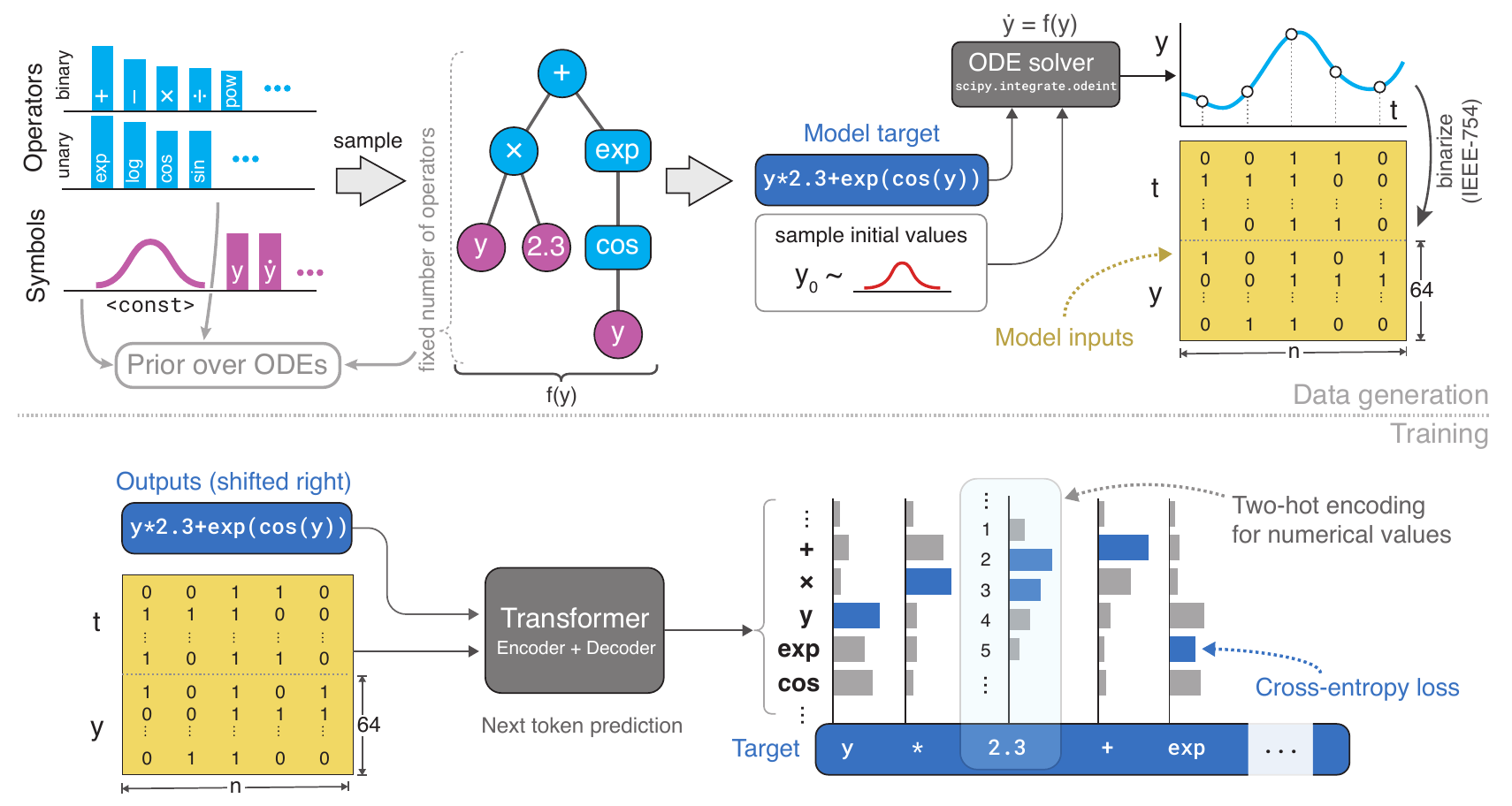}
  \caption{An overview illustration of the data generation (top) and training pipeline (bottom). Our dataset stores solutions in numerical (non-binarized) form on the entire regular solution time grid.}
  \label{fig:overview}
  \vspace{-2mm}
\end{figure*}

\section{Background and Related Work}
\label{sec:related_work}
While NODE \citep{chen2018neural} (with a large body of follow up work) is perhaps the most prominent method to learn ODEs from data in black-box form, we focus on various works that infer governing laws in symbolic form.
Classically, symbolic regression aims at regular functional relationships (mapping $(x, f(x))$ pairs to~$f$ instead of mapping trajectories $(t, y(t))$ to the governing ODE $\dot{y}=f(y, t)$) and has been approached by heuristics-based search, most prominently via genetic programming \citep{koza}. Genetic programming randomly evolves a population of prospective mathematical expressions over many iterations and mimics natural selection by keeping only the best contenders across iterations, where superiority is measured by user-defined and problem-specific fitness functions.
More recently, symbolic regression has been approached with machine learning methods which exploit gradient information to optimize within the space of (finite) compositions of pre-defined basis functions.
\citet{sindy} use linear regression to identify a (sparse) linear combination of basis functions that yields the best fit for the observed data, while other approaches use neural networks with a diverse set of activation functions \citep{eql2, pdenet2, rodenet}.
All these techniques deploy strong sparsity-promoting regularizers and fit a separate model for each observed trajectory.

Alternatively, one can train a model to directly output the symbolic expressions.
Supervised learning with gradient-based optimization for this approach is challenged by the formulation of a differentiable loss that measures the fit between the predicted symbolic expression and the observed data.
Thus, prior work resorted to reinforcement learning \citep{deepsymres} or evolutionary algorithms \citep{atkinson2019data, costa2021fast} for gradient-free optimization.
Furthermore, inspired by common properties of known natural laws, \citet{aifeynman2} devise a hybrid approach that combines a gradient-free heuristic search with neural network-based optimization. This approach has been extended to work with dynamical systems by \citet{weilbach2021inferring}.

The closest works to ours use pre-trained, attention-based sequence-to-sequence models for symbolic regression \emph{of functional relationships} \citep{nesymres,SymbolicGPT2021,kamienny2022end, vastl2022symformer}.
They exploit the fact that symbolic expressions for (multi-variate) scalar functions can be both generated and evaluated on random inputs cheaply, resulting in essentially unlimited training data.
Large data including ground-truth expressions in symbolic form allow for a differentiable cross-entropy loss based directly on the symbols of the expression, instead of the numerical proximity between evaluations of the predicted and true expression.
While the cross-entropy loss works well for operators and symbols (e.g. \texttt{+,exp,sin,x,y}), a naive implementation is inefficient for numerical constants, e.g., \texttt{1.452}. Previous works therefore resort to one of two strategies: 
1) represent all constants with a special \texttt{<const>} token when training the sequence-to-sequence model and predict only the presence of a constant. Actual values are then inferred in a second, subsequent parameter estimation step where the structure of an expression is held fixed and only constants are optimized.
This second optimization procedure comes with substantial computational cost as constants have to be fit per inferred expression. In particular, we highlight that it does not transfer to inferring ODEs as it would require to first solve the predicted ODE $\dot{y} = \hat{f}(y)$ to obtain predicted $\{\hat{y}_i\}_{i=1}^n$ values that can be compared to the set of observations $\{y_i\}_{i=1}^n$. While differentiable ODE solvers exist, optimizing constants this way is prohibitively expensive and typically highly unstable. 
2) A second popular strategy consists in rounding constants within the range of interest so that they can be represented with a finite number of tokens. This second strategy avoids a subsequent optimization step and enjoys clever encoding schemes with improved token efficiency yet represents values with an inherent loss in precision.
As an alternative, we develop a representation based on a ``two-hot'' encoding which avoids subsequent optimization steps as well as rounding.

\section{Method}
\label{sec:method}

\xhdr{Problem setting}
Given observations $\{(t_i, y_i)\}_{i=1}^n$ of a trajectory $y: [t_1, t_n] \to \bR$ that is a solution of the ODE $\dot{y} = f(y)$, we aim to recover the function~$f$ in symbolic form---in our case as a string.
In this work, we focus on time-invariant (or autonomous) ODEs (i.e.,~$f(y, t) = f(y)$).
Such settings are a good starting point for investigation as they are commonly studied and can be thought of as ``evolving on their own'' without external driving forces or controls, i.e., once an initial condition is fixed the absolute time does not directly influence the evolution.
We explicitly assume that the observed system actually evolves according to an ODE in canonical form~$\dot{y} = f(y)$ such that~$f$ can be expressed in closed form using the mathematical operators seen during training (see \cref{sec:data}).
In this paper we restrict ourselves to the rich class of non-linear, scalar, first-order, autonomous ODEs but we discuss extensions of NSODE to higher-order systems of coupled non-autonomous ODEs in \cref{app:extensions}.

\subsection{Data Generation}
\label{sec:data}

\xhdr{Sampling symbolic expressions}
\label{xhdr:sampling}
To exploit large-scale supervised pretraining we generate a dataset of $\sim$63M ODEs in symbolic form along with numerical solutions for randomly sampled initial values.
Since we assume ODEs to be in canonical form $\dot{y} = f(y)$, generating an ODE is equivalent to generating a symbolic expression $f(y)$.
We follow \citet{lample2019deep}, who sample such an expression $f(y)$ as a unary-binary tree, where each internal node corresponds to an operator and each leaf node corresponds to a constant or variable.
The algorithm consists of two phases: (1) A unary-binary tree is sampled uniformly from the distribution of unary-binary trees with up to $k\in \bN$ internal nodes, which crucially does not overrepresent small trees corresponding to short expressions. Here the maximum number of internal nodes $\nnodes$ is a hyperparameter of the algorithm.
(2) The sampled tree is ``decorated'', that is, each binary internal node is assigned a binary operator, each unary internal node is assigned a unary operator, and each leaf is assigned a variable or constant. 
Hence, we have to specify a distribution over the $\nbinops$ binary operators, one over the $\nunops$ unary operators, a probability $\psymb \in (0,1)$ to decide between symbols and constants, as well as a distribution $\pconst$ over constants.
For constants we distinguish explicitly between sampling an integer or a real value.
Together with $\nnodes$, these choices uniquely determine a distribution over equations $f$ and are described in detail in \cref{app:modeltraining}.
\Cref{fig:overview} depicts an overview of the data generation procedure.

The pre-order traversal of a sampled tree results in the symbolic expression for~$f$ in prefix notation.
After conversion to infix notation, we simplify each expression using the computer algebra system SymPy \citep{sympy}, and filter out constant equations~$f(y) = c$ as well as expressions that contain operators or symbols that were not part of the original distribution. 
We call the structure modulo the value of the constants of such an expression a \textbf{skeleton}.
Any skeleton containing at least one binary operator or constant can be represented by different unary-binary trees.
Vice versa many of the generated trees will be simplified to the same skeleton. To ensure diversity and to mitigate potential dataset bias towards particular expressions, we discard duplicates on the skeleton level. To further cheaply increase the variability of ODEs we sample $\nconsts$ unique sets of constants per skeleton.
When sampling constants we take care not to modify the canonical expression by adhering to the rules listed in \cref{app:constant_rules}.
We provide summary statistics on operator frequencies and expression complexities for the generated dataset in \cref{app:datastats}. Here, \textbf{complexity} refers to overall count of symbols (e.g., $y$, or constants) as well as operators in an expression, a simple yet common measure in the symbolic regression literature.

\xhdr{Computing numerical solutions}
We obtain numerical solutions for all ODEs via SciPy's interface \citep{scipy} to the LSODA software package \citep{lsoda} with both relative and absolute tolerances set to $10^{-9}$.
We solve each equation on a fixed time interval $t \in [0, \maxtime]$ and store solutions on a regular grid of $\ngrid$ points.
For each ODE, we sample up to $\nivs$ initial values $y(0) = y_0$ uniformly from $\ivrange$.\footnote{Due to a timeout per ODE, fewer solutions may remain in cases when the numerical solver fails repeatedly.}
While LSODA attempts to select an appropriate solver, numerical solutions still cannot be trusted in all cases.
Therefore, we check the validity of solutions via the following quality control check: we use 9th order central finite differences to approximate the temporal derivative of the solution trajectory (on the same temporal grid as the proposed solution), denoted by $\dot{y}_{\mathrm{fd}}$, and filter out any solution for which $\|\dot{y}_{\mathrm{fd}} - \dot{y} \|_{\infty} > \epsilon$, where we use $\epsilon = 1$.

\subsection{Model Design Choices}
\label{sec:model}

NSODE consists of an encoder-decoder transformer with architecture choices listed in \cref{app:modeldesign}.
We provide a visual overview in \cref{fig:overview}.

\xhdr{Representing input trajectories}
A key difficulty in feeding numerical solutions $\{y_i\}_{i=1}^n$ as input is that their range may differ greatly both within a single solution as well as across ODEs.
For example, the linear ODE $\dot{y} = c \cdot y$ for a constant~$c$ is solved by an exponential $y(t) = y_0 \exp(c t)$ for initial value $y(0) = y_0$, which may span many orders of magnitude on a fixed time interval.
To prevent numerical errors and vanishing or exploding gradients caused by the large range of values, we assume each representable 64-bit float value is a token and use its IEEE-754 encoding as the token representation \citep{nesymres}.
We thus convert all pairs $(t_i, y_i)$ to their IEEE-754 64 bit representations, channel them through a linear embedding layer before feeding them to the encoder.

\xhdr{Representing symbolic expressions}
The target sequence (i.e., the string for the symbolic expression of~$f$) is tokenized on the symbol-level.
We distinguish two cases: (1) \emph{Operators and variables:} for each operator and variable we include a unique token in the vocabulary. 
(2) \emph{Numerical constants:} constants may come from both discrete (integers) as well as continuous distributions, as for example in \texttt{y**2+1.64*cos(y)}. Hence, it is unfeasible to include individual tokens ``for each constant''.
Naively tokenizing on the digit level, i.e., representing real values literally as the sequence of characters (e.g., \texttt{"1.64"}), not only significantly expands the length of target sequences and thus the computational cost, but also requires a variable number of prediction steps for every single constant.

Instead, we take inspiration from \citet{schrittwieser2020mastering} and encode constants in a \emph{two-hot} fashion.
We fix a finite homogeneous grid on the real numbers $x_1 < x_2 < \ldots < x_m$ for some~$m \in \bN$, which we add as tokens to the vocabulary.
The grid range $(x_1, x_m)$ and number of grid points $m$ are hyperparameters that can be set in accordance to the problems of interest. Our choices are described in \cref{app:modeldesign}.

For any constant $c$ in the target sequence we then find $i \in \{1, \ldots, m-1\}$ such that $x_i \le c \le x_{i+1}$ and encode~$c$ as a distribution supported on $x_i, x_{i+1}$ with weights $\alpha, \beta$ such that $\alpha x_i + \beta x_{i+1} = c$.
That is, the target in the cross-entropy loss for a constant token is not a strict one-hot encoding, but a distribution supported on two (neighboring) vocabulary tokens resulting in a lossless encoding of continuous values in $[x_1, x_m]$.
This two-hot representation can be used directly in the cross-entropy loss function.

\xhdr{Decoding constants}
When decoding a predicted sequence, we check at each prediction step whether the argmax of the logits corresponds to one of the~$m$ constant tokens $\{x_1, \ldots, x_m\}$.
If not, we proceed by conventional one-hot decoding to obtain predicted operators and variables.
If instead the argmax corresponds to, for example, $x_i$, we also pick its largest-logit neighbor ($x_{i-1}$ or $x_{i+1}$; suppose $x_{i+1}$), renormalize their probabilities by applying a softmax to all logits and use the resulting two probability estimates as weights $\alpha, \beta$.
Constants are then ultimately decoded as $\alpha x_i + \beta x_{i+1}$.

\subsection{Evaluation and Metrics}
\label{sec:metrics}

\xhdr{Sampling solutions}
To infer a symbolic expression for the governing ODE of a new observed solution trajectory $\{(t_i, y_i)\}_{i=1}^n$, all the typical policies such as greedy, sampling, or beam search are available.
In our evaluation, we use beam search with 1536 beams and report top-$k$ results with $k$ ranging from 1 to 1536.

\xhdr{Metrics}
We evaluate model performance both numerically and symbolically.
For numerical evaluation we follow \citet{nesymres}: suppose the ground truth ODE is given by $\dot{y} = f(y)$ with (numerical) solution $y(t)$ and the predicted ODE is given by $\hat{\dot{y}} = \hat{f}(y)$.
To compute numerical accuracy we first evaluate $f$ and $\hat{f}$ on $\neval$ points in the interval $[\min(y(t)), \max(y(t))]$ (i.e., the interval traced out by the observed solution), which yields function evaluations $\texttt{gt}=\{\dot{y}_i\}_{i=1}^{\neval}$ and $\texttt{pred}=\{\hat{\dot{y}}_i\}_{i=1}^{\neval}$.
We then assess whether \texttt{numpy.allclose}\footnote{\texttt{numpy.allclose} returns True if \texttt{abs(a - b) <= (atol + rtol * abs(b))} holds element-wise for elements $a$ and $b$ from the two input arrays.
We use \texttt{atol=1e-10} and \texttt{rtol=0.05}; $a$ corresponds to predictions, $b$ corresponds to ground truth.} returns \texttt{True} as well as whether the coefficient of determination $\mathrm{R}^2 \geq 0.999$.\footnote{For observations $y_i$ and predictions $\hat{y}_i$ we have $\mathrm{R}^2 = 1 - (\sum_i (y_i - \hat{y}_i)^2) / (\sum_i (y_i - \overline{y})^2 )$.}
Numerical evaluations capture how closely the predicted function approximates the ground truth function within the interval $[\min(y(t)), \max(y(t))]$.

However, a key motivation for symbolic regression is to uncover a \emph{symbolic} mathematical expression that governs the observations.
Testing for symbolic equivalence between ground truth expression $f(y)$ and a predicted expression $\hat{f}(y)$ is unsuitable in the presence of real-valued constants as even minor deviations between true and predicted constants render the equivalence false.
Instead, we regard the predicted expression $\hat{f}(y)$ to be symbolically correct if $f(y)$ and $\hat{f}(y)$ can be made equivalent by modifying only the values of constants in the predicted expression $\hat{f}(y)$.
This is implemented using SymPy's \texttt{match} function.
In order not to alter the structure of the predicted expression, we constrain modifications of constants such that all constants remain non-zero and retain their original sign.
This definition is thus primarily concerned with the structure of an expression, rather than precise numerical agreement.
Once the structure is known, the inference problem becomes conventional parameter estimation.
We report percentages of samples in a given test set that satisfies any individual metric (numerical and symbolic), as well as percentages satisfying symbolic and numerical metrics simultaneously.

\section{Experiments}
\label{sec:experiments}

\subsection{Benchmark Datasets}\label{sec:datasets}
We construct several test sets to evaluate model performance and generalization in different settings.
\begin{itemize}[leftmargin=*,topsep=0pt,itemsep=0pt]
    \item \textbf{testset-iv}: Our first test set assesses generalization within initial values not seen during training.
    It consists of 5793 ODEs picked uniformly at random from our generated dataset but re-sampled initial values. We also employ the following constraints via rejection sampling: (a) All skeletons in testset-iv are unique. (b) As the number of unique skeletons increases with the number of operators, we allow at most 2000 examples per number of operators (with substantially fewer unique skeletons existing for few operators).
    
    \item \textbf{testset-constants}: Our second test set assesses generalization within unseen initial values and constants.
    It consists of 2720 ODEs picked uniformly at random from our dataset (ensuring unique skeletons and at most 1000 examples per number of operators as above), but re-sampled intial values and constants.
    
    \item \textbf{testset-skeletons}: In principle, we can train NSODE on all possible expressions (using only the specified operators and number ranges) up to a specified number of operators.
    However, even with the millions examples in our dataset, we have by far not exhausted the huge space of possible skeletons (especially for larger numbers of operators).
    Hence, our third test set contains 100 novel random ODEs with skeletons that were never seen during training. 
    
    \item \textbf{testset-iv-163}: This is a subset of testset-iv motivated by the fact that most symbolic regression models we want to compare to require a separate optimization for every individual example, which was computationally infeasible for our testset-iv.
    For a fair comparison, we therefore subsampled up to 10 ODEs per complexity uniformly at random, yielding 163 examples.
    
    \item \textbf{testset-textbook}: To assess how NSODE performs on ``real problems'', we manually curated 12 scalar, non-linear ODEs from Wikipedia pages, physics textbooks, and lecture note from university courses on ODEs.
    These equations are listed in \cref{tab:textbook_equations} in \cref{app:textbook_equations}.
    We note that they are all extremely simple compared to the expressions in our generated dataset in that they are ultimately mostly low order polynomials, some of which with one fractional exponent.
    
    \item \textbf{testset-classic}:
    To validate our approach on existing datasets we turn to benchmarks in the classic symbolic regression literature (inferring just the functional relationship between input-ouput pairs) and simply interpret functions as ODEs. In particular we include all scalar function listed in the overview in \cite{mcdermott2012genetic} which includes equations from many different benchmarks \cite{keijzer2003improving, koza, koza1994genetic, uy2011semantically, vladislavleva2008order}.
    For example, we interpret the function $f(y) = y^3 + y^2 + y$ from \citet{uy2011semantically} as an autonomous ODE $\dot{y}(t) = f(y(t)) = y(t)^3 + y(t)^2 + y$, which we solve numerically for a randomly sampled initial value as described before.
\end{itemize}

\subsection{Baselines}\label{sec:baselines}
We compare our method to recent popular baselines from the literature (see \cref{sec:related_work}).
We briefly describe them including some limitations here and defer all details to \cref{app:baselines}.
First, no baseline is suited directly to infer dynamical laws, but only to infer functional relationships.
Therefore, all baselines fit a separate regression function mapping $y(t) \mapsto \hat{\dot{y}}(t)$ per individual ODE, using the coefficient of determination $\mathrm{R}^2$ as optimization objective.
Since derivatives $\hat{\dot{y}}(t)$ are typically not observed, we approximate them via finite differences using PySindy \citep{pysindy}.
Hence, all these methods crucially rely on regularly sampled and noise-free observations, whereas our approach can easily be extended to take those into account (see \cref{app:extensions}).

\begin{itemize}[leftmargin=*,topsep=0pt,itemsep=0pt]
    \item \textbf{Sindy} \citep{sindy}: Sindy builds a (sparse) linear combination of a fixed set of (non-linear) basis functions.
    The resulting Lasso regression is efficient, but suffers from limited expressiveness.
    In particular, Sindy cannot easly represent nested functions or non-integer powers as all non-linear expressions have to be added explicitly to the set of basis functions.
    We cross-validate Sindy over a fairly extensive hyperparameter grid of 800 different combinations for each individual trajectory.
    
    \item \textbf{GPL}\footnote{\ttfamily \url{gplearn.readthedocs.io/}} (genetic programming):
    GPL(earn) maintains a population of programs each representing a mathematical expression.
    The programs are mutated for several generations to heuristically optimize a user defined fitness function.
    While not originally developed for ODEs, we can apply GPLearn on our datasets by leveraging the finite difference approximation.
    We use a population size of 1000 and report the best performance across all final programs.
    Compared to sindy, GPLearn is more expressive yet substantially slower to fit.
    
    \item \textbf{AIFeynman} \citep{aifeynman, aifeynman2}: AIFeynman is a physics-inspired approach to symbolic regression that exploits the insight that many famous equations in natural sciences exhibit well-understood functional properties such as symmetries, compositionality, or smoothness.
    AIFeynman implements a neural network based heuristic search that tests for such properties in order to identify a symbolic expression that fits the data.
    For every test sample AIFeynman computes a pareto front of solutions that trade off complexity versus accuracy.
    We report the best performance across all functions on the pareto front. 
    Notably, AIFeynman performs quite an exhaustive search procedure such that running it even on a single equation took on the order of tens of minutes.
\end{itemize}

\subsection{Results}\label{sec:results}

\begin{figure}
\centering
\begin{subfigure}{1\textwidth}
    \centering
    \includegraphics[width=0.65\linewidth]{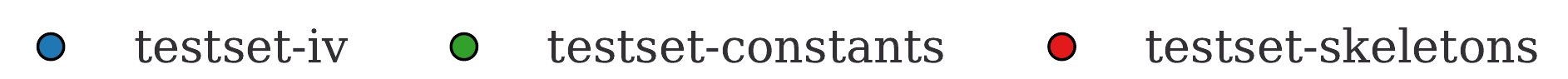}
\end{subfigure}%
\\ \vspace{0.25cm}
\begin{subfigure}{.25\textwidth}
    \centering
    \includegraphics[width=1\linewidth]{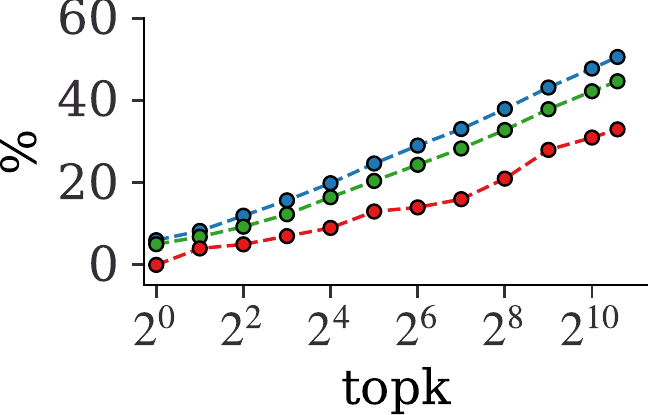}
    \caption{allclose}
\end{subfigure}%
\begin{subfigure}{.25\textwidth}
    \centering
    \includegraphics[width=1\linewidth]{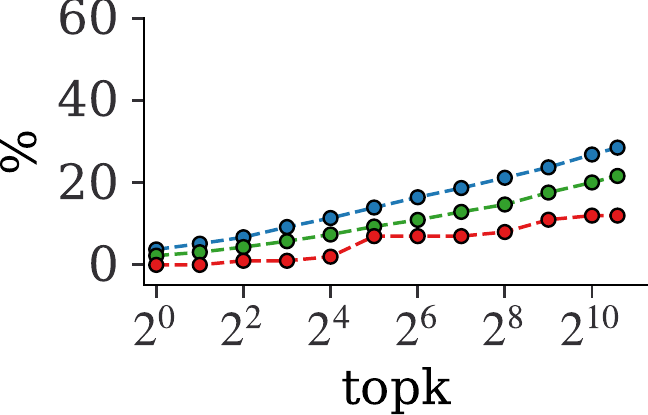}
    \caption{R$^2 \geq 0.999$}
\end{subfigure}%
\begin{subfigure}{.25\textwidth}
    \centering
    \includegraphics[width=1\linewidth]{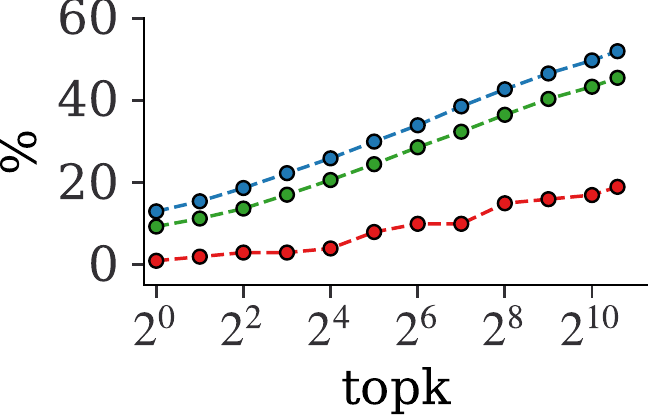}
    \caption{skeleton recovery}
\end{subfigure}%
\begin{subfigure}{.25\textwidth}
    \centering
    \includegraphics[width=1\linewidth]{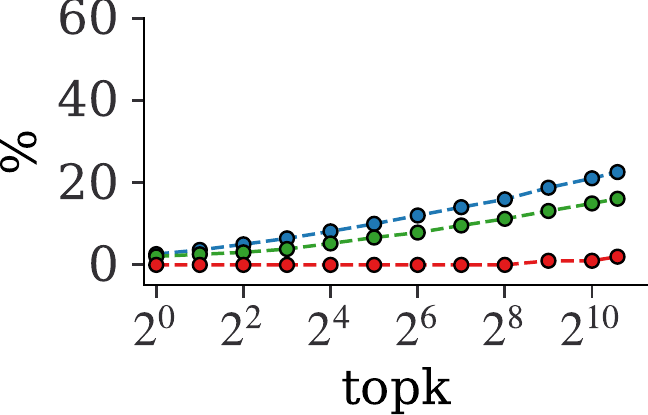}
    \caption{skel. recov. \& allclose}
\end{subfigure}%
\caption{Numerical and symbolic performance evaluation on testset-iv.}
\label{fig:results}
\end{figure}

\xhdr{Model Performance}
\Cref{fig:results} shows NSODE's performance on our testset-iv, testset-constants, and testset-skeletons according to our numerical and symbolic metrics as well as combined skeleton recovery and allclose as we vary $k$ in the top-k considered candidates of the beam search.
Investing sufficient test-time-compute (i.e., considering many candidates) continuously improves performance.
While we capped $k$ at 1536 due to memory limitations, we did not observe a stagnation of the roughly logarithmic scaling of all performance metrics with $k$.
This cannot be attributed to ``exhaustion effects'', where one may assume that all possible ODEs will eventually be among the candidates, because (a) the space of possible skeletons grows much faster than exponentially, and (b) the numerical metrics are extremely sensitive also to the predicted constant values in continuous domains.

As one may expect, performance decreases as we move from only new initial conditions, to also sampling new constants, to finally sampling entirely unseen skeletons.
On testset-iv with $k=1536$ we achieve about 50\% skeleton recovery and still successfully recover more than a third skeletons of testset-skeletons with similar numbers for allclose.
The fact that the combined metric (symbolic + numerical) is only about half of that indicates that numerical and symbolic fit are indeed two separate measures, none of which need to imply the other.
Hence, a thorough evaluation of both is crucial to understand model performance in symbolic regression tasks.

\begin{table}[h!]
\centering
\caption{Comparing NSODE to the baselines. Results are average percentages across dataset. GPLearn often generates extremely long expressions which take SymPy up to half a minute to parse during evaluation. We denote this extra time in gray.}\label{tab:resultsummary}
\begin{tabularx}{\columnwidth}{l@{\hspace{8pt}}l@{\hspace{8pt}}Y@{\hspace{6pt}}Y@{\hspace{8pt}}Y@{\hspace{0pt}}Y}
\toprule \rowcolor{white}
Dataset                           & Metric                 & NSODE & Sindy & GPLearn & AIFeynman  \\ 
\midrule
                                   & skel-recov             & \bf 37.4  & 3.7   & 2.5 &  14.1   \\ 
                                  & R$^2 \geq 0.999$              & 24.5  &  31.9  & 3.7& \bf 49.7     \\ 
iv-163                    & allclose               & 42.3  & 25.8  & 14.7 & \bf 55.8  \\ 
                                  & skel-recov \& R$^2 \geq 0.999$  & \bf 15.3  & 3.1   & 1.8 &  13.5   \\ 
                                  & skel-recov \& allclose & \bf 15.3  & 3.1   & 1.8  &  13.5  \\ 
                                  & runtime [s]             & 5.4 & \bf0.4 & 29 {\color{gray}+22} & 1203.6 \\
\midrule
                                    & skel-recov             & 41.7  & 33.3  & 8.3  & \bf 91.7   \\ 
                                  & R$^2 \geq 0.999$              &  16.7  & 50   & 0.0  & \bf 75   \\ 
textbook                            & allclose               & 25    &  58.3  & 8.3  & \bf 75   \\ 
                                  & skel-recov  \& R$^2 \geq 0.999$ &  33.3   &  41.7  & 0 & \bf 66.7      \\ 
                                  & skel-recov \& allclose & 8.3  &  33.3  & 1.8 & \bf 66.7   \\
                                  & runtime [s]             & 6 & \bf1 & 23 {\color{gray}+22} & 1267.1 \\
\bottomrule
\end{tabularx}%
\end{table}

\xhdr{Comparison to baselines}
In \cref{tab:resultsummary} we compare NSODE to all baselines using $k=1536$ in our beam search; full results on all datasets can be found in \cref{app:detailedresults}.
We also include the average wallclock runtime per expression for each of the datasets.

\begin{figure}
\centering
\begin{subfigure}{1\textwidth}
    \centering
    \includegraphics{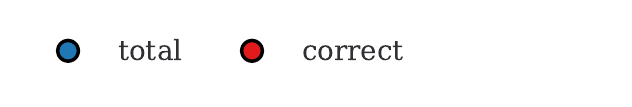}
\end{subfigure}%
\\
\begin{subfigure}{.25\textwidth}
    \centering
    \includegraphics[width=.99\linewidth]{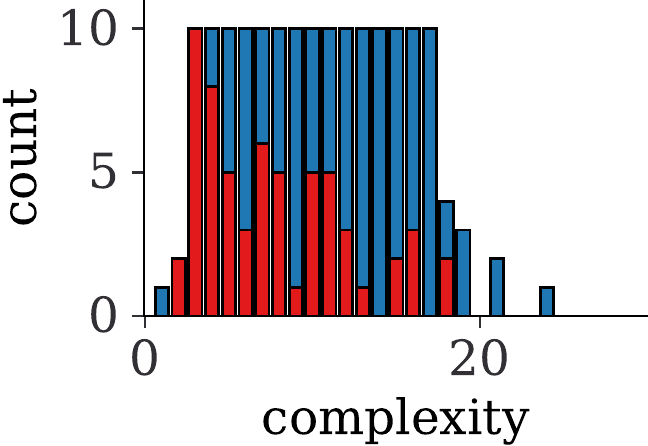}
    \caption{NSODE}
\end{subfigure}%
\begin{subfigure}{.25\textwidth}
    \centering
    \includegraphics[width=.99\linewidth]{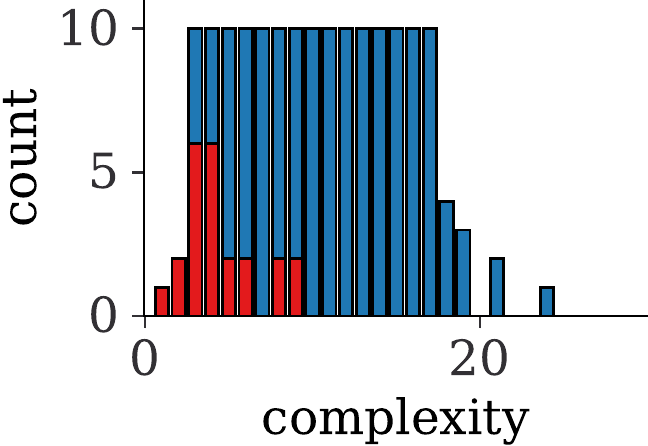}
    \caption{AIFeynman}
\end{subfigure}%
\begin{subfigure}{.25\textwidth}
    \centering
    \includegraphics[width=.99\linewidth]{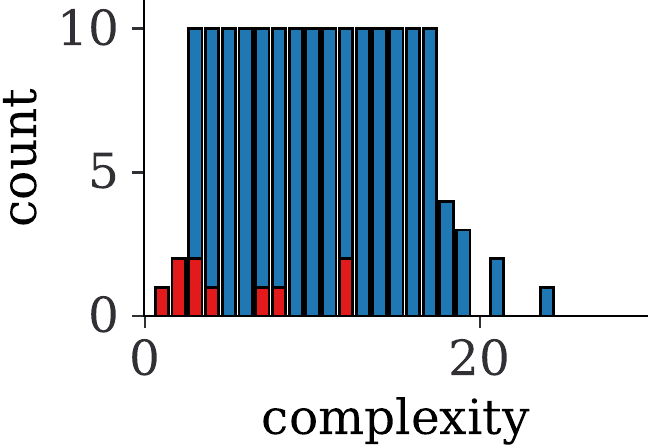}
    \caption{Sindy}
\end{subfigure}%
\begin{subfigure}{.25\textwidth}
    \centering
    \includegraphics[width=.99\linewidth]{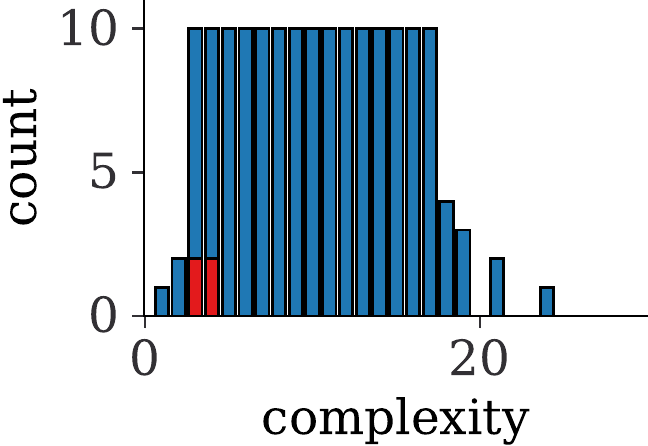}
    \caption{GPLearn}
\end{subfigure}%
\caption{Correctly recovered skeletons by each method on testset-iv-163 per complexity. AIFeynman and Sindy are mostly able to recover some of the low complexity skeletons, while NSODE performs much better also on higher complexities. GPLearn fails to recover most skeletons.}\label{fig:distribution}
\vspace{-2mm}
\end{figure}

First, we note that on our subsampled testset-iv-163, NSODE outperforms competing approaches in terms of skeleton recovery by a wide margin and also performs best in terms of joint skeleton recovery and numerical measures, which is a strong indication of actually having recovered the governing ODE accurately.
By spending over 200x more time on its exhaustive heuristic search, AIFeynman manages to outperform NSODE in terms of numerical accuracy ($\mathrm{R}^2$ and allclose).
\Cref{fig:distribution} shows the number of skeletons recovered by each method given the complexity of equations, results for other datasets can be found in \cref{app:detailedresults}. \footnote{Due to simplification, complexity is not upper bounded by the number of nodes in a unary-binary tree.}
While AIFeynman and Sindy recover some of the low complexity expressions, NSODE is the only method to also recover some of the more complex skeletons. 

On testset-textbook, AIFeynman outperforms all other methods on numerical and symbolic metrics. This can be understood with regard to the dataset where $8/12$ expressions are polynomials with the remaining 4/12 expressions having a polynomial skeleton with fractional or negative exponents. These expressions are particularly favorable for the heuristics implemented by AIFeynman which explicitly attempt to fit a polynomial to the data.
However, even on these simple examples AIFeynman takes over 200x longer than our method, which in turn clearly outperforms Sindy and GPLearn in terms of skeleton recovery.

\section{Conclusion}
\label{sec:conclusion}

We have developed a flexible and scalable method to infer ordinary differential equations $\dot{y} = f(y)$ from a single observed solution trajectory.
Our method follows the successful paradigm of large-scale pretraining of attention-based sequence-to-sequence models on essentially unlimited amounts of simulated data, where the inputs are the observed solution $\{(t_i,y_i)\}_{i=1}^n$ and the output is a symbolic expression for~$f$ as a string.
Once trained, our method is orders of magnitude more efficient than similarly expressive existing symbolic regression techniques that require a separate optimization for each instance and achieves strong performance in terms of skeleton recovery especially for complex expressions on various benchmarks.

\bibliographystyle{abbrvnat}
\bibliography{bibliography}

\begin{thebibliography}{28}
\providecommand{\natexlab}[1]{#1}
\providecommand{\url}[1]{\texttt{#1}}
\expandafter\ifx\csname urlstyle\endcsname\relax
  \providecommand{\doi}[1]{doi: #1}\else
  \providecommand{\doi}{doi: \begingroup \urlstyle{rm}\Url}\fi

\bibitem[Atkinson et~al.(2019)Atkinson, Subber, Wang, Khan, Hawi, and
  Ghanem]{atkinson2019data}
S.~Atkinson, W.~Subber, L.~Wang, G.~Khan, P.~Hawi, and R.~Ghanem.
\newblock Data-driven discovery of free-form governing differential equations.
\newblock \emph{arXiv preprint arXiv:1910.05117}, 2019.

\bibitem[Biggio et~al.(2021)Biggio, Bendinelli, Neitz, Lucchi, and
  Parascandolo]{nesymres}
L.~Biggio, T.~Bendinelli, A.~Neitz, A.~Lucchi, and G.~Parascandolo.
\newblock Neural symbolic regression that scales.
\newblock In M.~Meila and T.~Zhang, editors, \emph{Proceedings of the 38th
  International Conference on Machine Learning}, volume 139 of
  \emph{Proceedings of Machine Learning Research}, pages 936--945. PMLR, 18--24
  Jul 2021.
\newblock URL \url{https://proceedings.mlr.press/v139/biggio21a.html}.

\bibitem[Brunton et~al.(2016)Brunton, Proctor, and Kutz]{sindy}
S.~L. Brunton, J.~L. Proctor, and J.~N. Kutz.
\newblock Discovering governing equations from data by sparse identification of
  nonlinear dynamical systems.
\newblock \emph{Proceedings of the National Academy of Sciences}, 113\penalty0
  (15):\penalty0 3932--3937, 2016.
\newblock ISSN 0027-8424.
\newblock \doi{10.1073/pnas.1517384113}.
\newblock URL \url{https://www.pnas.org/content/113/15/3932}.

\bibitem[Chen et~al.(2018)Chen, Rubanova, Bettencourt, and
  Duvenaud]{chen2018neural}
R.~T. Chen, Y.~Rubanova, J.~Bettencourt, and D.~Duvenaud.
\newblock Neural ordinary differential equations.
\newblock In \emph{Proceedings of the 32nd International Conference on Neural
  Information Processing Systems}, pages 6572--6583, 2018.

\bibitem[Costa et~al.(2021)Costa, Dangovski, Dugan, Kim, Goyal, Soljačić, and
  Jacobson]{costa2021fast}
A.~Costa, R.~Dangovski, O.~Dugan, S.~Kim, P.~Goyal, M.~Soljačić, and
  J.~Jacobson.
\newblock Fast neural models for symbolic regression at scale, 2021.

\bibitem[de~Silva et~al.(2020)de~Silva, Champion, Quade, Loiseau, Kutz, and
  Brunton]{pysindy}
B.~de~Silva, K.~Champion, M.~Quade, J.-C. Loiseau, J.~N. Kutz, and S.~Brunton.
\newblock Pysindy: A python package for the sparse identification of nonlinear
  dynamical systems from data.
\newblock \emph{Journal of Open Source Software}, 5\penalty0 (49):\penalty0
  1--4, 2020.

\bibitem[Hindmarsh and Laboratory(1982)]{lsoda}
A.~Hindmarsh and L.~L. Laboratory.
\newblock \emph{ODEPACK, a Systematized Collection of ODE Solvers}.
\newblock Lawrence Livermore National Laboratory, 1982.
\newblock URL \url{https://books.google.de/books?id=9XWPmwEACAAJ}.

\bibitem[Kamienny et~al.(2022)Kamienny, d'Ascoli, Lample, and
  Charton]{kamienny2022end}
P.-A. Kamienny, S.~d'Ascoli, G.~Lample, and F.~Charton.
\newblock End-to-end symbolic regression with transformers.
\newblock \emph{arXiv preprint arXiv:2204.10532}, 2022.

\bibitem[Keijzer(2003)]{keijzer2003improving}
M.~Keijzer.
\newblock Improving symbolic regression with interval arithmetic and linear
  scaling.
\newblock In \emph{European Conference on Genetic Programming}, pages 70--82.
  Springer, 2003.

\bibitem[Koza(1993)]{koza}
J.~R. Koza.
\newblock \emph{Genetic programming - on the programming of computers by means
  of natural selection}.
\newblock Complex adaptive systems. {MIT} Press, 1993.
\newblock ISBN 978-0-262-11170-6.

\bibitem[Koza(1994)]{koza1994genetic}
J.~R. Koza.
\newblock \emph{Genetic programming II: automatic discovery of reusable
  programs}.
\newblock MIT press, 1994.

\bibitem[Lample and Charton(2019)]{lample2019deep}
G.~Lample and F.~Charton.
\newblock Deep learning for symbolic mathematics.
\newblock In \emph{International Conference on Learning Representations}, 2019.

\bibitem[Liu et~al.(2020)Liu, Long, Wang, Sun, and Dong]{rodenet}
J.~Liu, Z.~Long, R.~Wang, J.~Sun, and B.~Dong.
\newblock Rode-net: learning ordinary differential equations with randomness
  from data.
\newblock \emph{arXiv preprint arXiv:2006.02377}, 2020.

\bibitem[Long et~al.(2019)Long, Lu, and Dong]{pdenet2}
Z.~Long, Y.~Lu, and B.~Dong.
\newblock Pde-net 2.0: Learning pdes from data with a numeric-symbolic hybrid
  deep network.
\newblock \emph{Journal of Computational Physics}, 399:\penalty0 108925, 2019.

\bibitem[McDermott et~al.(2012)McDermott, White, Luke, Manzoni, Castelli,
  Vanneschi, Jaskowski, Krawiec, Harper, De~Jong, et~al.]{mcdermott2012genetic}
J.~McDermott, D.~R. White, S.~Luke, L.~Manzoni, M.~Castelli, L.~Vanneschi,
  W.~Jaskowski, K.~Krawiec, R.~Harper, K.~De~Jong, et~al.
\newblock Genetic programming needs better benchmarks.
\newblock In \emph{Proceedings of the 14th annual conference on Genetic and
  evolutionary computation}, pages 791--798, 2012.

\bibitem[Meurer et~al.(2017)Meurer, Smith, Paprocki, \v{C}ert\'{i}k, Kirpichev,
  Rocklin, Kumar, Ivanov, Moore, Singh, Rathnayake, Vig, Granger, Muller,
  Bonazzi, Gupta, Vats, Johansson, Pedregosa, Curry, Terrel, Rou\v{c}ka, Saboo,
  Fernando, Kulal, Cimrman, and Scopatz]{sympy}
A.~Meurer, C.~P. Smith, M.~Paprocki, O.~\v{C}ert\'{i}k, S.~B. Kirpichev,
  M.~Rocklin, A.~Kumar, S.~Ivanov, J.~K. Moore, S.~Singh, T.~Rathnayake,
  S.~Vig, B.~E. Granger, R.~P. Muller, F.~Bonazzi, H.~Gupta, S.~Vats,
  F.~Johansson, F.~Pedregosa, M.~J. Curry, A.~R. Terrel, v.~Rou\v{c}ka,
  A.~Saboo, I.~Fernando, S.~Kulal, R.~Cimrman, and A.~Scopatz.
\newblock Sympy: symbolic computing in python.
\newblock \emph{PeerJ Computer Science}, 3:\penalty0 e103, Jan. 2017.
\newblock ISSN 2376-5992.
\newblock \doi{10.7717/peerj-cs.103}.
\newblock URL \url{https://doi.org/10.7717/peerj-cs.103}.

\bibitem[Petersen et~al.(2021)Petersen, Landajuela, Mundhenk, Santiago, Kim,
  and Kim]{deepsymres}
B.~K. Petersen, M.~Landajuela, T.~N. Mundhenk, C.~P. Santiago, S.~K. Kim, and
  J.~T. Kim.
\newblock Deep symbolic regression: Recovering mathematical expressions from
  data via risk-seeking policy gradients.
\newblock In \emph{Proc. of the International Conference on Learning
  Representations}, 2021.

\bibitem[Sahoo et~al.(2018)Sahoo, Lampert, and Martius]{eql2}
S.~Sahoo, C.~Lampert, and G.~Martius.
\newblock Learning equations for extrapolation and control.
\newblock In \emph{International Conference on Machine Learning}, pages
  4442--4450. PMLR, 2018.

\bibitem[Schrittwieser et~al.(2020)Schrittwieser, Antonoglou, Hubert, Simonyan,
  Sifre, Schmitt, Guez, Lockhart, Hassabis, Graepel,
  et~al.]{schrittwieser2020mastering}
J.~Schrittwieser, I.~Antonoglou, T.~Hubert, K.~Simonyan, L.~Sifre, S.~Schmitt,
  A.~Guez, E.~Lockhart, D.~Hassabis, T.~Graepel, et~al.
\newblock Mastering atari, go, chess and shogi by planning with a learned
  model.
\newblock \emph{Nature}, 588\penalty0 (7839):\penalty0 604--609, 2020.

\bibitem[Udrescu and Tegmark(2020)]{aifeynman}
S.-M. Udrescu and M.~Tegmark.
\newblock Ai feynman: A physics-inspired method for symbolic regression.
\newblock \emph{Science Advances}, 6\penalty0 (16):\penalty0 eaay2631, 2020.

\bibitem[Udrescu et~al.(2020)Udrescu, Tan, Feng, Neto, Wu, and
  Tegmark]{aifeynman2}
S.-M. Udrescu, A.~Tan, J.~Feng, O.~Neto, T.~Wu, and M.~Tegmark.
\newblock Ai feynman 2.0: Pareto-optimal symbolic regression exploiting graph
  modularity.
\newblock \emph{arXiv preprint arXiv:2006.10782}, 2020.

\bibitem[Uy et~al.(2011)Uy, Hoai, O’Neill, McKay, and
  Galv{\'a}n-L{\'o}pez]{uy2011semantically}
N.~Q. Uy, N.~X. Hoai, M.~O’Neill, R.~I. McKay, and E.~Galv{\'a}n-L{\'o}pez.
\newblock Semantically-based crossover in genetic programming: application to
  real-valued symbolic regression.
\newblock \emph{Genetic Programming and Evolvable Machines}, 12\penalty0
  (2):\penalty0 91--119, 2011.

\bibitem[Valipour et~al.(2021)Valipour, Panju, You, and
  Ghodsi]{SymbolicGPT2021}
M.~Valipour, M.~Panju, B.~You, and A.~Ghodsi.
\newblock {SymbolicGPT: A Generative Transformer Model for Symbolic
  Regression}.
\newblock In \emph{Preprint Arxiv}, 2021.
\newblock URL \url{https://arxiv.org/abs/2106.14131}.

\bibitem[Vastl et~al.(2022)Vastl, Kulh{\'a}nek, Kubal{\'i}k, Derner, and
  Babu{\v{s}}ka]{vastl2022symformer}
M.~Vastl, J.~Kulh{\'a}nek, J.~Kubal{\'i}k, E.~Derner, and R.~Babu{\v{s}}ka.
\newblock Symformer: End-to-end symbolic regression using transformer-based
  architecture.
\newblock \emph{arXiv preprint arXiv:2205.15764}, 2022.

\bibitem[Virtanen et~al.(2020)Virtanen, Gommers, Oliphant, Haberland, Reddy,
  Cournapeau, Burovski, Peterson, Weckesser, Bright, {van der Walt}, Brett,
  Wilson, Millman, Mayorov, Nelson, Jones, Kern, Larson, Carey, Polat, Feng,
  Moore, {VanderPlas}, Laxalde, Perktold, Cimrman, Henriksen, Quintero, Harris,
  Archibald, Ribeiro, Pedregosa, {van Mulbregt}, and {SciPy 1.0
  Contributors}]{scipy}
P.~Virtanen, R.~Gommers, T.~E. Oliphant, M.~Haberland, T.~Reddy, D.~Cournapeau,
  E.~Burovski, P.~Peterson, W.~Weckesser, J.~Bright, S.~J. {van der Walt},
  M.~Brett, J.~Wilson, K.~J. Millman, N.~Mayorov, A.~R.~J. Nelson, E.~Jones,
  R.~Kern, E.~Larson, C.~J. Carey, {\.I}.~Polat, Y.~Feng, E.~W. Moore,
  J.~{VanderPlas}, D.~Laxalde, J.~Perktold, R.~Cimrman, I.~Henriksen, E.~A.
  Quintero, C.~R. Harris, A.~M. Archibald, A.~H. Ribeiro, F.~Pedregosa, P.~{van
  Mulbregt}, and {SciPy 1.0 Contributors}.
\newblock {{SciPy} 1.0: Fundamental Algorithms for Scientific Computing in
  Python}.
\newblock \emph{Nature Methods}, 17:\penalty0 261--272, 2020.
\newblock \doi{10.1038/s41592-019-0686-2}.

\bibitem[Vladislavleva et~al.(2008)Vladislavleva, Smits, and
  Den~Hertog]{vladislavleva2008order}
E.~J. Vladislavleva, G.~F. Smits, and D.~Den~Hertog.
\newblock Order of nonlinearity as a complexity measure for models generated by
  symbolic regression via pareto genetic programming.
\newblock \emph{IEEE Transactions on Evolutionary Computation}, 13\penalty0
  (2):\penalty0 333--349, 2008.

\bibitem[Weilbach et~al.(2021)Weilbach, Gerwinn, Weilbach, and
  Kandemir]{weilbach2021inferring}
J.~Weilbach, S.~Gerwinn, C.~Weilbach, and M.~Kandemir.
\newblock Inferring the structure of ordinary differential equations.
\newblock \emph{arXiv preprint arXiv:2107.07345}, 2021.

\bibitem[Zaheer et~al.(2020)Zaheer, Guruganesh, Dubey, Ainslie, Alberti,
  Ontanon, Pham, Ravula, Wang, Yang, et~al.]{zaheer2020big}
M.~Zaheer, G.~Guruganesh, K.~A. Dubey, J.~Ainslie, C.~Alberti, S.~Ontanon,
  P.~Pham, A.~Ravula, Q.~Wang, L.~Yang, et~al.
\newblock Big bird: Transformers for longer sequences.
\newblock \emph{Advances in Neural Information Processing Systems},
  33:\penalty0 17283--17297, 2020.

\end{thebibliography}

\newpage
\appendix

\section{Possible extensions}
\label{app:extensions}

While we have focused exclusively on the huge class of scalar, autonomous, first-order ODEs, we believe that our approach can scale also to non-autonomous, higher-order, systems of ODEs.

\xhdr{Non-autonomous equations} Since our model is provided with time $t$ as inputs, it is capable of learning functions $f(y, t)$ depending on $y$ and $t$ explicitly.
Hence, extending our approach to non-autonomous ODEs is simply a matter of adding a symbol for $t$ in our data generation.

\xhdr{Irregular samples and noise} Due to the separate data generation and training phase, it is straight forward to train NSODE on corrupted input sequences, where we could for example add observation noise (need not be additive) to the $\{y_i\}_{i=1}^n$ or randomly drop some of the observations on the regular observation grid to simulate irregularly sampled observations.
We expect these models to still perform well and in particular much better than the baselines, since their targets are approximated derivatives, which are highly sensitive to noise and irregular samples.

\xhdr{Systems of equations} For a system of $K$ first-order ODEs, the data generation requires two updates: first, we have to generate $K$ equations $(f_k)_{k \in \{1,\ldots, K\}}$, one for each component. In addition, each of those functions can depend not only on $y$  (and $t$ in the non-autonomous case), but $K$ different components $\{y^{(k)}\}_{k \in \{1, \ldots, K\}}$. This is easily achieved by allowing for $K$ tokens $y^{(k)}$ in the data generation.
We can then simply augment the input sequence to the transformer to contain not only $(t, y)$ as for the scalar case, but $(t, y^{(1)}, y^{(2)}, \ldots, y^{(K)})$ for a system of $K$ variables.
Finally, we have decide on a way to set the target for the transformer, i.e., how to represent the system of equations in symbolic form as \emph{a single sequence}.
A straight forward way to achieve this is to introduce a special delimitation character, e.g., \texttt{"|"} to separate the different components. Note that the order in which to predict the $K$ equations is dictated by the order in which they are stacked for the input sequence, hence this information is in principle available to the model.
For example, in the case of a two-dimensional system, where $y = (y^{(1)}, y^{(2)}) \in \bR^2$, we could have the model perform the following mapping
\begin{equation*}
  \{(t_i, y^{(1)}_i, y^{(2)}_i)\}_{i=1}^n \to \texttt{"-1.4*y\textsuperscript{(1)}*y\textsuperscript{(2)}+cos(t) | sin(y\textsuperscript{(1)}+y\textsuperscript{(2)})"} \:,
\end{equation*}
where the separator \texttt{|} delimits the two components of~$f(y,t)$ corresponding to~$\dot{y}^{(1)}$ and $\dot{y}^{(2)}$.

\xhdr{Higher-order equations} It is well known that any higher order (system of) ODEs can be reduced to a first-order system of ODEs. Specifically, a $d$-th order system of $K$ variables can be reduced to an equivalent first-order system of $d\cdot K$ variables.
Hence, one can handle higher-order systems analogously as before with multiple separator tokens.
One obstacle in this case is that when only observations of $y(t)$ are given, one first needs to obtain observed derivatives to reduce a higher-order system to a first-order system.
These would in turn have to be estimated from data, which suffers from the same challenges we have mentioned previously (instability under noise and irregular samples).

Finally, when we want to have a single model deal with higher-order equations of unknown order, or systems with differing numbers of variables, it remains on open question how to have the model automatically adjust to the potentially vastly differing input dimensions or how to automatically detect the order of an ODE.


\section{Implementation Details}
\label{app:modeltraining}

\subsection{Rules to Resample Constants}
\label{app:constant_rules}
As described in \cref{xhdr:sampling}, we generate ODEs as unary-binary trees, convert them into infix notation and parse them into a canonical form using \texttt{sympy}. From each skeleton we then create up to 25 ODEs by sampling different values for the constants. When resampling constants we want to ensure that we do not accidentally modify the skeleton as this would additionally burden our model with resolving potential ambiguities in the grammar of ODE expressions. Furthermore, we do not want to reintroduce duplicate samples on the skeleton level after carefully filtering them out previously. We therefore introduce the following sampling rules for constants:
\begin{enumerate}
    \item Do not sample constants of value 0.
    \item When the original constant in the skeleton is negative, sample a negative constant, otherwise sample a positive constant.
    \item Avoid base of 1 in power operations as $1^x = 1$.
    \item Avoid exponent of 1 and -1 in power operations as $x^1 = x$ and $x^{-1} = 1/x$.
    \item Avoid coefficients of value 1 and -1 as $1 \cdot x = x$ and $-1 \cdot x = -x$
    \item Avoid divisions by 1 and -1 as $x/1 = x$ and $x/-1 = -x$
\end{enumerate}

\subsection{Data generation}
\label{app:datageneration}

As discussed in the main text, the choices of the maximum number of internal nodes per tree $\nnodes$, the choice and distribution over $\nbinops$ binary operators, the choice and distribution over $\nunops$ unary operators, the probability with which to decorate a leaf with a symbol $\psymb$ (versus a constant with $1-\psymb$), and the distribution $\pconst$ over constants uniquely determine the training distribution over ODEs~$f$.
These choices can be viewed as flexible and semantically interpretable tuning knobs to choose a prior over ODEs.
For example, it may be known in a given context, that the system follows a ``simple'' law (small $\nnodes$) and does not contain exponential rates of change (do not include $\exp$ in the unary operators), and so on.
The choice of the maximum number of operators per tree, how to sample the operators, and how to fill in the leaf nodes define the training distribution, providing us with flexible and semantically meaningful tuning knobs to choose a prior over ODE systems for our model.
We summarize our choices in \cref{tab:dataparams,tab:binary_operators,tab:unary_operators}, where $\mathcal{U}$ denotes the uniform distribution.
Whenever a leaf node is decorated with a constant, the distribution over constants is determined by first determining whether to use an integer or a real value with equal probablity.
In case of an integer, we sample it from $\pint$, and in case of a real-valued constant we sample it from $\preal$ shown in \cref{tab:dataparams}. Finally, when it comes to the numerical solutions of the sampled ODEs, we fixed the parameters in \cref{tab:solutionparams} for our experiments.

\begin{table}[b!]
    \caption{Parameter settings for the data generation.}\label{tab:dataparams}
    \centering
    \begin{tabularx}{\columnwidth}{@{}lccccYc@{}}
    \rowcolor{white}
    \toprule
    parameter  & $\nnodes$ & $\nbinops$ & $\nunops$ & $\psymb$ & $\pint$ & $\preal$ \\ \midrule
    value      & 5 & 5 & 5 & 0.5 & $\mathcal{U}(\{-10, \ldots, 10\} \setminus \{0\})$ & $\mathcal{U}((-10, 10))$ \\
    \bottomrule
    \end{tabularx}%
\end{table}

\begin{table}[b!]
    \caption{Binary operators with their relative sampling frequencies}\label{tab:binary_operators}
    \centering
    \begin{tabularx}{0.75\columnwidth}{l*5{Y}}
    \rowcolor{white}
    \toprule
    operator  & $+$ & $-$ & $\cdot$ & $\div$ & pow \\ \midrule
    \rowcolor{white}
    probability & 0.2 & 0.2 & 0.2     & 0.2    & 0.2\\
    \bottomrule
    \end{tabularx}%
\end{table}

\begin{table}[b!]
    \caption{Unary operators with their relative sampling frequencies. The last column denotes the unary minus as in e.g. $f(y) = -y$. We do not explicitly sample this operator but keep it after simplification.}\label{tab:unary_operators}
    \centering
    \begin{tabularx}{0.75\columnwidth}{l*6{Y}}
    \rowcolor{white}
    \toprule
    operator  & $\sin$ & $\cos$ & $\exp$ & $\sqrt{\phantom{x}}$ & $\log$ & $-$\\ \midrule
    probability & 0.2    & 0.2    & 0.2    & 0.2     & 0.2 & 0\\
    \bottomrule
    \end{tabularx}%
\end{table}

\begin{table}[b!]
    \caption{Parameters for numerical solutions of sampled ODEs.}\label{tab:solutionparams}
    \centering
    \begin{tabularx}{0.85\columnwidth}{l*6{Y}}
    \rowcolor{white}
    \toprule
    parameter  & $\nconsts$ & $\nivs$ & $\maxtime$ & $\ngrid$ & $\ivrange$  \\ \midrule
    \rowcolor{white}
    value & 25    & 25    & 4    & 1024     & $(-5,5)$ \\
    \bottomrule
    \end{tabularx}%
\end{table}

We highlight that there is no such thing as ``a natural distribution over equations'' when it comes to ODEs.
Hence, ad-hoc choices have to be made in one way or another.
However, it is important to note that neither our chosen range of integers nor the range of real values for constants are in any way restrictive as they can be achieved by appropriate rescaling.
In particular, the model itself represents these constant values merely be non-numeric tokens and interpolates between those anchor tokens (our two-hot encoding) to represent continuous values.
Hence, the model is entirely agnostic to the actual numerical range spanned by these fixed grid tokens, but the relative accuracy in recovering interpolated values will be constant and thus scale with the absolute chosen range.
Therefore, scaling $\pint$ and $\preal$ by essentially any power of 10 does not affect our findings.
Similarly, the chosen range of initial values $\ivrange$ is non-restrictive as one could simply scale each observed trajectory to have its starting value lie within this range.

\subsection{Model}
\label{app:modeldesign}

\xhdr{Architectural choices}
We use an encoder-decoder transformer and list our architectural choices in \cref{app:architecture}.
For the two-hot representation of symbolic constants as described in \cref{sec:model}, we choose an equidistant grid $-10=x_1 < x_2 < \ldots < x_m = 10$ with $m=21$.

While not relevant for our dataset (as we check for convergence of the ODE solvers), we remark that the input-encoding via IEEE-754 binary representations also graciously represents special values such as \texttt{nan} or \texttt{inf} without causing errors.
Those are thus valid inputs that may still provide useful training signal, e.g., ``the solution of the ODE of interest goes to \texttt{inf} quickly''.

\begin{table}
\centering
\vspace{-7mm}
\small
\caption{Overview of our model architecture.}\label{app:architecture}
\begin{tabularx}{\columnwidth}{l@{\hspace{6pt}}Y@{\hspace{0pt}}Y}
\toprule \rowcolor{white}
    & \textbf{Encoder} & \textbf{Decoder} \\ \midrule
    \textbf{architecture} & BigBird${}^{\dagger}$ & BigBird \\
    \textbf{layers} & 6 & 6 \\
    \textbf{heads} & 16 & 16  \\
    \textbf{embed. dim.} & 512 & 512 \\
    \textbf{forward dim.} & 2048 & 2048 \\
    \textbf{activation} & gelu & gelu  \\
    \textbf{vocab. size} & - &  43 \\
    \textbf{position enc.} & learned & learned \\
    \textbf{parameters} & 23.3M & 23.3M \\ \bottomrule
\end{tabularx}\\%
{\footnotesize${}^{\dagger}$We chose the BigBird \citep{zaheer2020big} implementation with full attention available in Hugging Face.}
\end{table}

\xhdr{Training details}
 Our model is trained on 4 Nvidia A100 GPUs for 18 epochs after which we evaluate the best model based on the validation loss. We choose a batchsize of 600 samples and use a linear learning rate warm-up over 10,000 optimization step after which we keep the learning rate constant at $10^{-4}$. During training we use teacher forcing to guide sequential predictions. Teacher forcing is straightforward to implement for one-hot encoded tokens (operators, symbols) which are simply passed through a learnable embedding layer before their embedded representations are fed to the decoder. In contrast, two-hot encoded tokens (constants) can not be passed through an embedding layer directly. For a generic constant in the target sequence represented as $\alpha x_i + \beta x_{i+1}$, we thus instead provide the linear combination of the two embeddings $\alpha\; \text{\texttt{embed(x\textsubscript{i})}} + \beta\; \text{\texttt{embed(x\textsubscript{i+1})}}$ as decoder input. We represent symbolic input to the model in prefix format which relieves the model from correctly predicting opening and closing parentheses.

\xhdr{Evaluation}
We use $\neval = 100$ for the evaluation of our numerical metrics.


\section{Dataset statistics}\label{app:datastats}
\begin{figure}[H]
\centering
  \begin{subfigure}{1\textwidth}
    \centering
    \includegraphics[width=\linewidth]{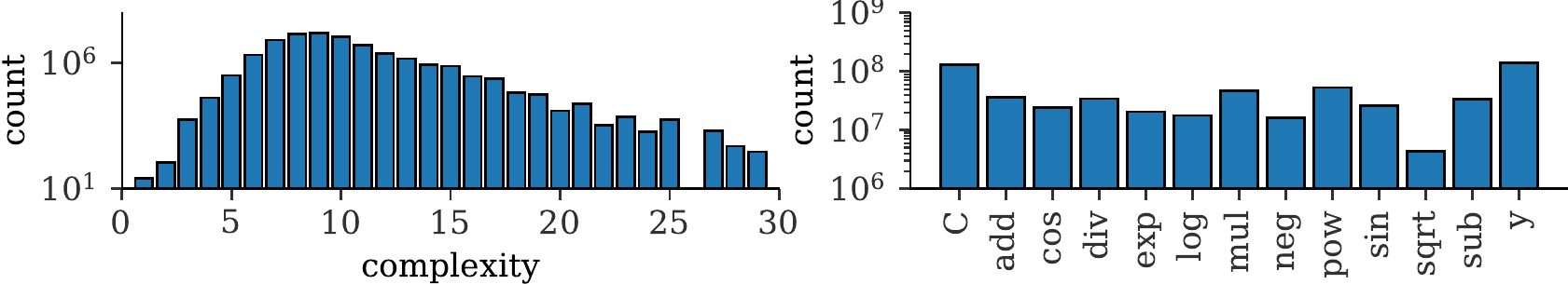}
    \caption{Training data}
  \end{subfigure}\\
  \begin{subfigure}{1\textwidth}
    \centering
    \includegraphics[width=\linewidth]{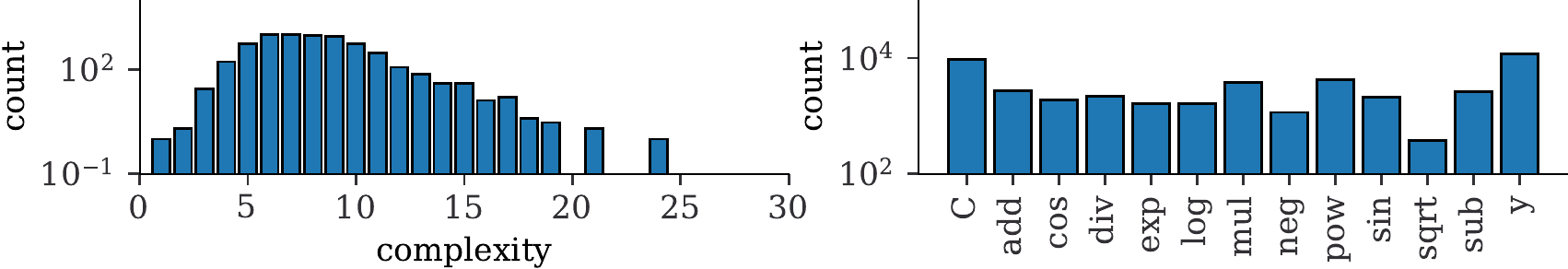}
    \caption{Testset-iv}
  \end{subfigure}\\
  \begin{subfigure}{1\textwidth}
    \centering
    \includegraphics[width=\linewidth]{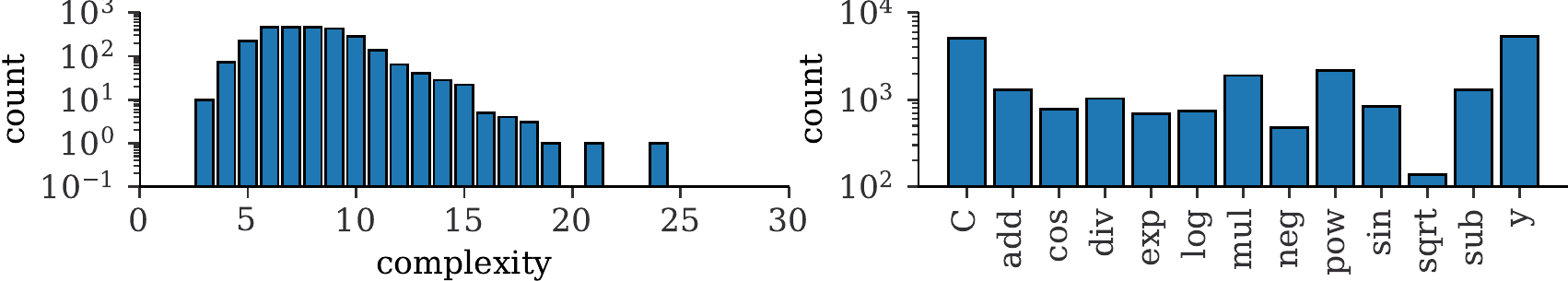}
    \caption{testset-const}
  \end{subfigure}\\
  \begin{subfigure}{1\textwidth}
    \centering
    \includegraphics[width=\linewidth]{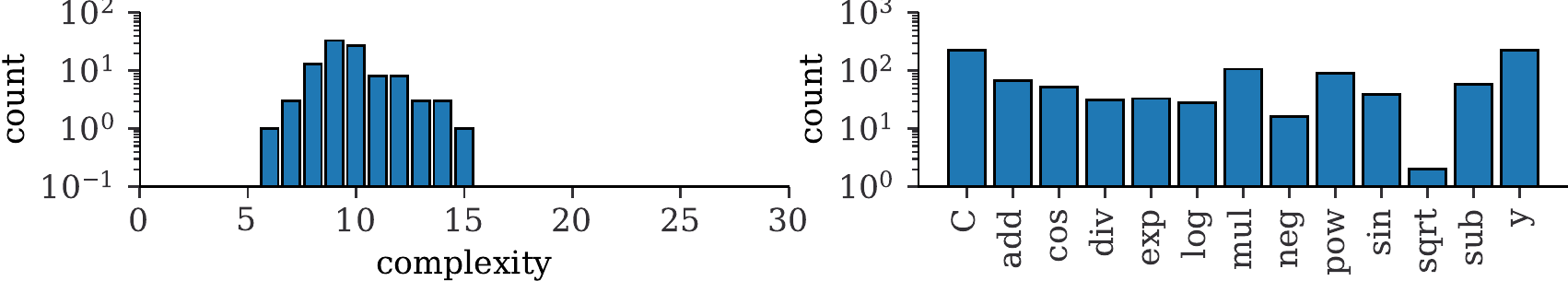}
    \caption{testset-skeleton}
  \end{subfigure}\\
  \begin{subfigure}{1\textwidth}
    \centering
    \includegraphics[width=\linewidth]{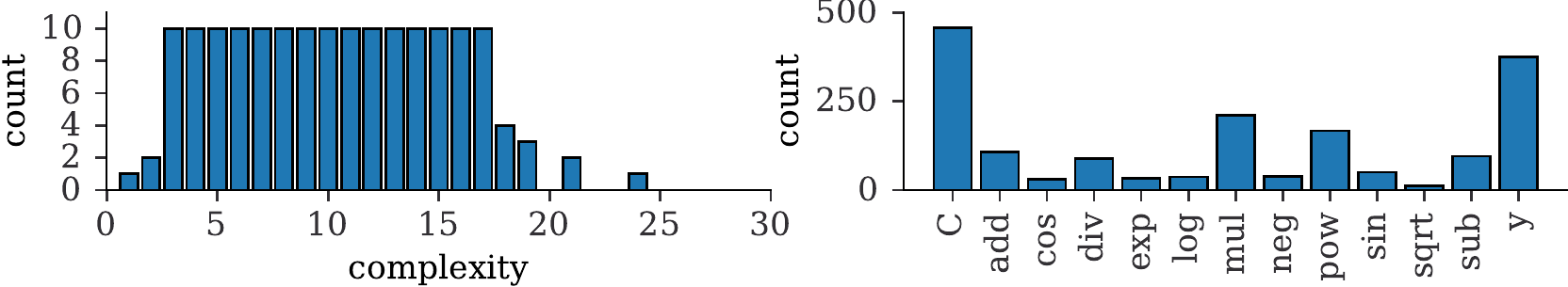}
    \caption{testset-iv-163}
  \end{subfigure}\\
  \begin{subfigure}{1\textwidth}
    \centering
    \includegraphics[width=\linewidth]{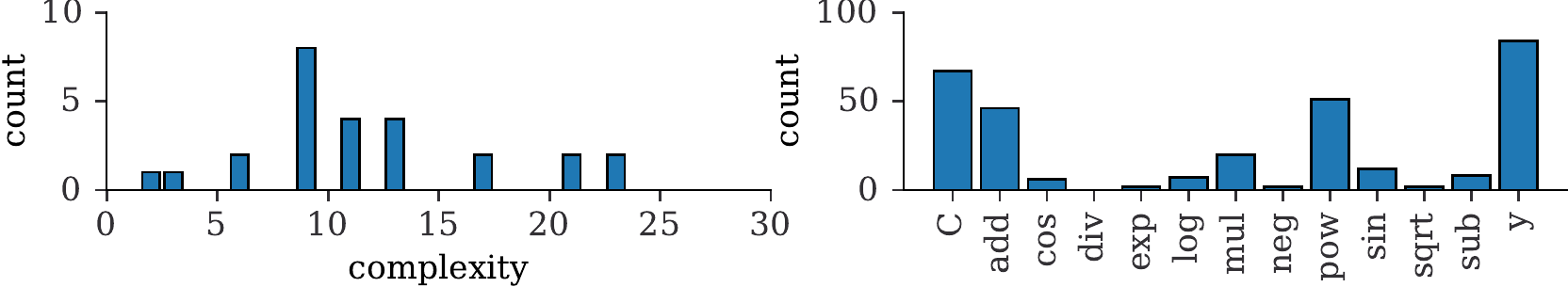}
    \caption{testset-classic}
  \end{subfigure}\\
  \begin{subfigure}{1\textwidth}
    \centering
    \includegraphics[width=\linewidth]{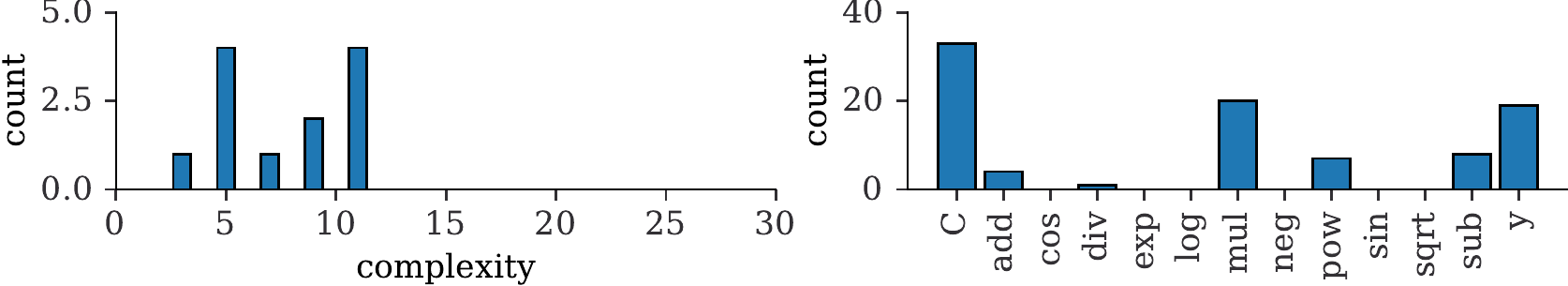}
    \caption{testset-textbook}
  \end{subfigure}
  \caption{Distribution of complexity and operators for all datasets. Complexity is defined as the number of symbols, constants and operators in an expression.\label{fig:datastats}}
\end{figure}

We provide an overview over the complexity distribution and the absolute frequency of all operators (after simplification) for all datasets in \cref{fig:datastats}. We can see that our self-generated dataset covers by far the larges complexity whereas both complexities and operator diversity are much lower for equations in the classic and textbook ODEs.

\section{Textbook equations dataset}
\label{app:textbook_equations}
\Cref{tab:textbook_equations} list the equations we collected from wikipedia, textbooks and lecture notes together with the initial values that we solved them for. We can also see that almost all of these equations simplify to low-order polynomials.

\begin{table}
\centering
\caption{Equations of the textbook dataset.\label{tab:textbook_equations}}

\begin{tabularx}{\columnwidth}{lYYr}
\toprule \rowcolor{white}
\textbf{Name} & \textbf{Equation} $f(x)$ & \textbf{simplified} &  $\mathbf{y_0}$ \\ \midrule
autonomous Riccati                                                                     & $0.6\cdot y^2+2\cdot y+0.1$                                                         & $0.6\cdot y^2+2\cdot y+0.1$      & $-0.2$          \\\hline
autonomous Stuart-Landau                                                               & $-2.2/2\cdot y^3 + 1.31\cdot y$                                                      & $-1.1\cdot y^3 + 1.31\cdot y$          & $0.1$           \\\hline
autonomous Bernoulli                                                                   & $-1.3\cdot y+2.1\cdot y^{2.2}$                                                       & $-1.3\cdot y + 2.1\cdot y^{2.2}$         & $0.6$           \\\hline
compound interest                                                                      & $0.1\cdot y$                                                                         & $0.1\cdot y$                       & $9$             \\\hline
Newton's law of cooling                                                                & $-0.1\cdot(y-3)$                                                                     & $0.3-0.1\cdot y$                 & $9$             \\\hline
Logistic equation                                                                      & $0.23\cdot y\cdot (1-y)$                                                             & $0.23\cdot(y-y^2)$            & $9$             \\\hline
\begin{tabular}[c]{@{}l@{}}Logistic equation \\ with harvesting\end{tabular}           & $0.23\cdot y\cdot (1-0.33\cdot y) - 0.5$                                             & $0.23\cdot y-0.76\cdot y^2-0.5$ & $9$             \\\hline
\begin{tabular}[c]{@{}l@{}}Logistic equation \\ with harvesting 2\end{tabular}         & $2\cdot y\cdot(1-y/3) - 0.5$                                                         & $2\cdot y-0.66\cdot y^2-0.5$ & $0.7$           \\\hline
Solow-Swan                                                                             & \begin{tabular}[c]{@{}l@{}}$y^0.5\cdot (0.9\cdot  8 - (3 + 2.5)\cdot$\\$y^{1- 0.5})$ \end{tabular}                              & $7.2\cdot y^{0.5}-5.5\cdot y$   & $0.1$           \\\hline
Tank draining                                                                          & $-\sqrt{2\cdot 9.81}\cdot (2/9)^2\cdot\sqrt{y}$                                       & $-0.21\cdot y^{0.5}$                & $1$             \\\hline
\begin{tabular}[c]{@{}l@{}}Draining water \\ through a funnel\end{tabular}             & \begin{tabular}[c]{@{}l@{}}$-(0.5^2/4)\cdot\sqrt{2\cdot 9.81}\cdot$\\ $(\sin{1}/\cos{1})^2\cdot y^{-1.5}$\end{tabular} & $-0.67/y^{1.5}$                & $3$             \\\hline
\begin{tabular}[c]{@{}l@{}}velocity of a body\\ thrown vertically upwards\end{tabular} & $-9.81 - 0.9\cdot  y/8.2$                                             & $-0.10\cdot y - 9.81$              & $0.1$           \\ 
\bottomrule
\end{tabularx}
\end{table}


\section{Baselines}
\label{app:baselines}
We here describe more detail on the optimization of the baseline comparison models.

\xhdr{Sindy} We use the implementation available in PySindy \citep{pysindy} and instantiate the basis functions with polynomials up to degree 10 as well as all unary operators listed in \cref{tab:unary_operators}. When fitting sindy to data we often encountered numerical issues especially when using high-degree polynomial or the exponential function. To attenuate such issues we set the highest degree of the polynomials per sample to the highest degree present in the ground truth. Secondly, when numerical issues are caused by a particular basis function, we discard this basis function for the current sample and restart the fitting process. We run a separate full grid search for every ODE over the following hyper-parameters and respective values (these all include the default values): 
\begin{itemize}[leftmargin=*,topsep=0pt,itemsep=0pt]
    \item optimizer-threshold (\texttt{np.logspace(-5, 0, 10)}): Minimum magnitude for a coefficient in the weight vector to not be zeroed out.
    \item optimizer-alpha ($[0.001, 0.0025, 0.005, 0.01, 0.025, 0.05, 0.1, 0.2]$): L2 regularizer on parameters.
    \item finite differences order ($[2, 3, 5, 7, 9]$): Order of finite difference approximation.
    \item maximum number of optimization iterations ([20, 100]): Maximum number of optimization steps.
\end{itemize}
For every ODE, sindy is fit using solution trajectory in the initial interval $[0, 2]$ and validated on the interval $(2, 4]$. The grid search thus results in a ranking of models with different hyper-parameter configurations. Instead of evaluating only the performance of the best model, we report top-k performance across the ranked hyper parameter configurations. Sindy is computationally highly efficient yet limited in its expressiveness, in particular it can not represent nested functions or non-integer powers.

\xhdr{GPLearn} We instantiate GPLearn with a constant range of $(-10, 10)$ and all binary operators listed in \cref{tab:binary_operators} and all unary operators listed \cref{tab:unary_operators} except for the exponential function which caused numerical issues. We keep the default hyper-parameters but run a grid search across the parsimony coefficient (\{0.0005, 0.001, 0.01, 0.05, 0.1, 0.2, 0.5, "auto"\}) which trades of fitness versus program length. We choose $R^2$ as fitness function.

\xhdr{AIFeynman} We use the AIFeynman implementation from \url{https://github.com/SJ001/AI-Feynman} and run the algorithm with the following (default) hyper-parameters:
\begin{itemize}
    \item Brute force try time: 60 seconds
    \item Number of epochs for the training : 500
    \item Operator set: 14
    \item Maximum degree of polynomial tried by the polynomial fit routine: 4
\end{itemize}

\section{Detailed results}\label{app:detailedresults}

\begin{figure}
\centering
\begin{subfigure}{1\textwidth}
    \centering
    \includegraphics[width=0.5\linewidth]{figs/complexity_vs_recovery/legend2.pdf}
\end{subfigure}\\
\begin{subfigure}{0.25\textwidth}
    \centering
    \includegraphics[width=\linewidth]{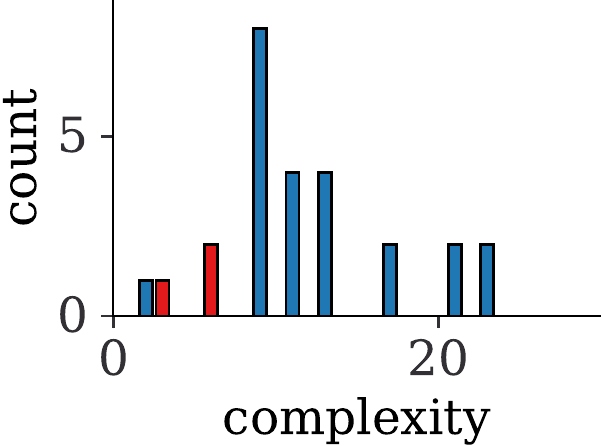}
    \caption{NSODE}
\end{subfigure}%
\begin{subfigure}{0.25\textwidth}
    \centering
    \includegraphics[width=\linewidth]{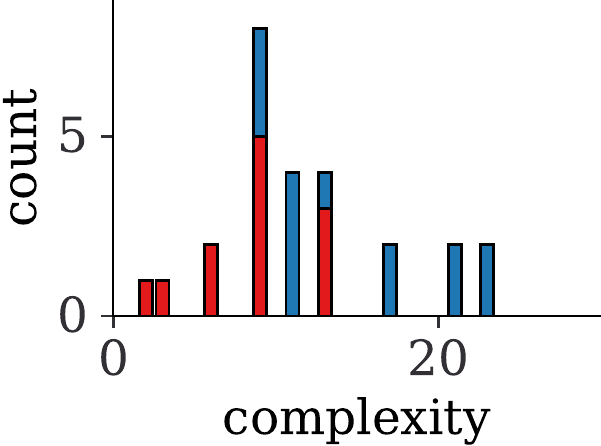}
    \caption{AIFeynman}
\end{subfigure}%
\begin{subfigure}{0.25\textwidth}
    \centering
    \includegraphics[width=\linewidth]{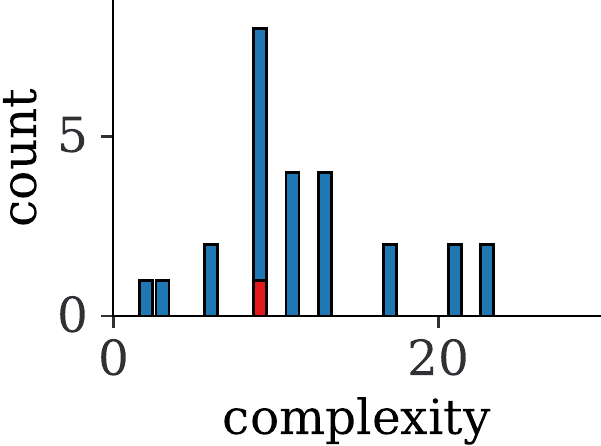}
    \caption{Sindy}
\end{subfigure}%
\begin{subfigure}{0.25\textwidth}
    \centering
    \includegraphics[width=\linewidth]{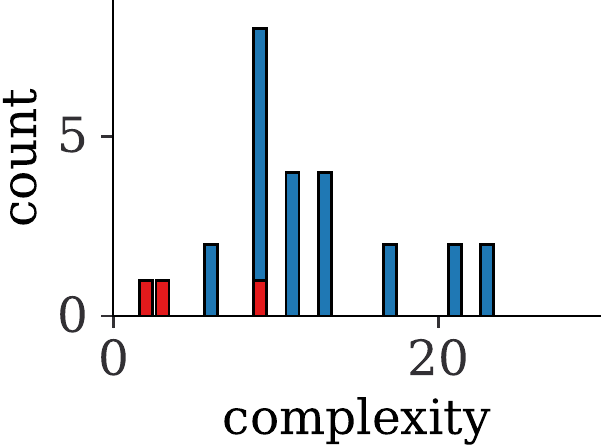}
    \caption{GPLearn}
\end{subfigure}
\caption{Correctly recovered skeletons by each method on the classic benchmark dataset per complexity.}\label{fig:classiccorrect}
\end{figure}

\begin{figure}
\centering     
\begin{subfigure}{1\textwidth}
    \centering
    \includegraphics[width=0.5\linewidth]{figs/complexity_vs_recovery/legend2.pdf}
\end{subfigure}\\
\begin{subfigure}{0.25\textwidth}
    \centering
    \includegraphics[width=\linewidth]{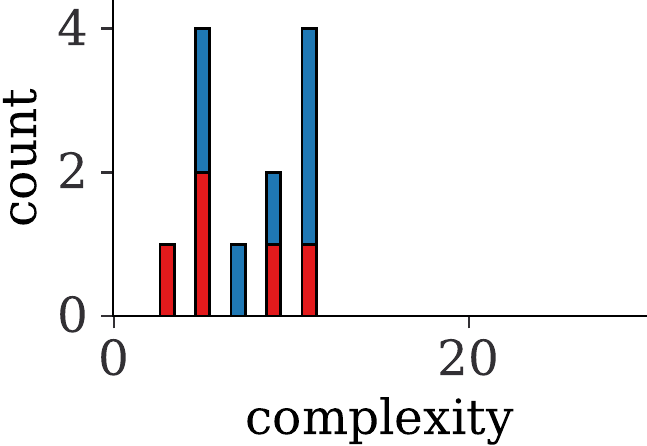}
    \caption{NSODE}
\end{subfigure}%
\begin{subfigure}{0.25\textwidth}
    \centering
    \includegraphics[width=\linewidth]{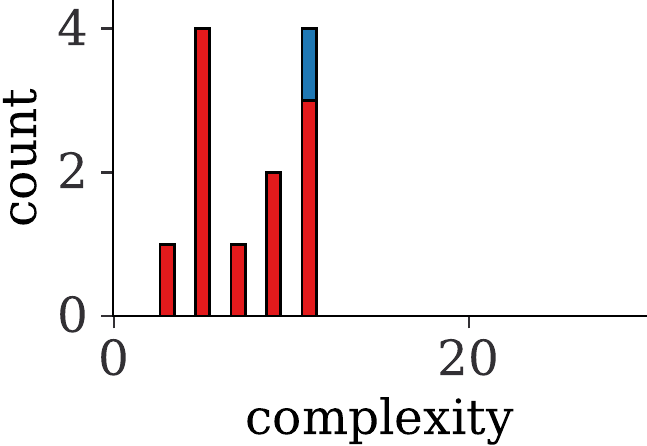}
    \caption{AIFeynman}
\end{subfigure}%
\begin{subfigure}{0.25\textwidth}
    \centering
    \includegraphics[width=\linewidth]{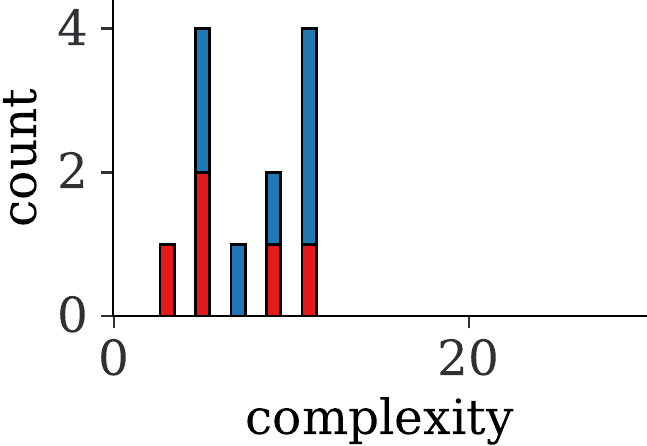}
    \caption{Sindy}
\end{subfigure}%
\begin{subfigure}{0.25\textwidth}
    \centering
    \includegraphics[width=\linewidth]{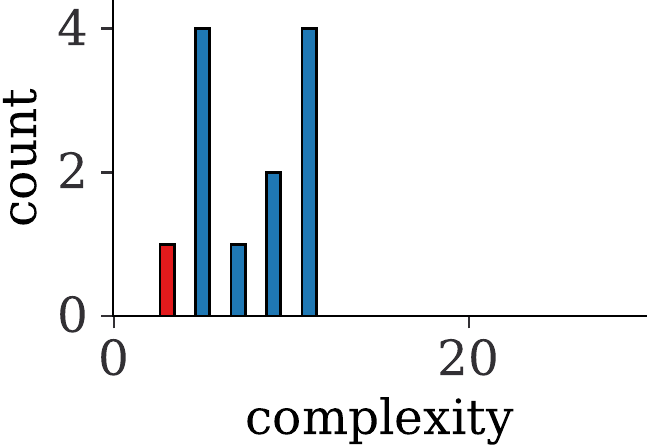}
    \caption{GPLearn}
\end{subfigure}
\caption{Correctly recovered skeletons by each method on the textbook dataset per complexity.}\label{fig:textbookcorrect}
\end{figure}

\begin{figure}
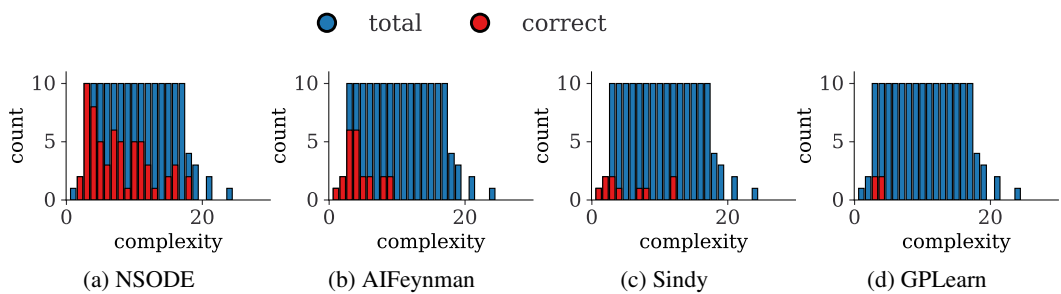

\centering
\begin{subfigure}{1\textwidth}
    \centering
    \includegraphics[width=0.5\linewidth]{figs/complexity_vs_recovery/legend2.pdf}
\end{subfigure}%
\\
\begin{subfigure}{.25\textwidth}
    \centering
    \includegraphics[width=\linewidth]{figs/complexity_vs_recovery/complexity_vs_recovery_nsode_163.pdf}
    \caption{NSODE}
\end{subfigure}%
\begin{subfigure}{.25\textwidth}
    \centering
    \includegraphics[width=\linewidth]{figs/complexity_vs_recovery/complexity_vs_recovery_aifeynman_163.pdf}
    \caption{AIFeynman}
\end{subfigure}%
\begin{subfigure}{.25\textwidth}
    \centering
    \includegraphics[width=\linewidth]{figs/complexity_vs_recovery/complexity_vs_recovery_sindy_163.pdf}
    \caption{Sindy}
\end{subfigure}%
\begin{subfigure}{.25\textwidth}
    \centering
    \includegraphics[width=\linewidth]{figs/complexity_vs_recovery/complexity_vs_recovery_gplearn_163.pdf}
    \caption{GPLearn}
\end{subfigure}%
\caption{Correctly recovered skeletons by each method on testset-iv-163 per complexity.}\label{fig:163correct}
\vspace{-2mm}
\end{figure}

\begin{figure}
\centering
\begin{subfigure}{1\textwidth}
    \centering
    \includegraphics[width=0.5\linewidth]{figs/complexity_vs_recovery/legend2.pdf}
\end{subfigure}%
\\
\begin{subfigure}{.25\textwidth}
    \centering
    \includegraphics[width=\linewidth]{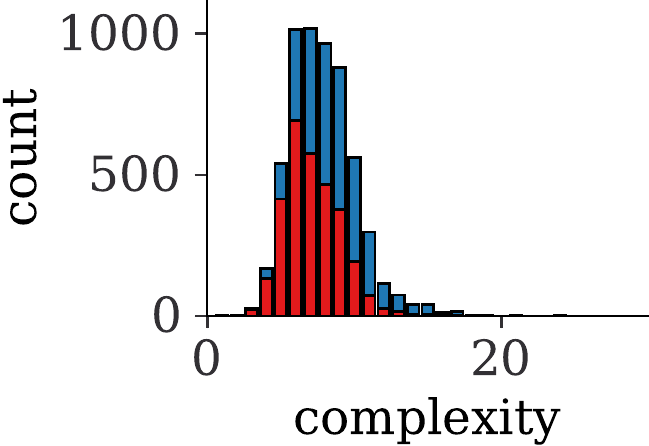}
    \caption{Testset-iv}
\end{subfigure}%
\begin{subfigure}{.25\textwidth}
    \centering
    \includegraphics[width=\linewidth]{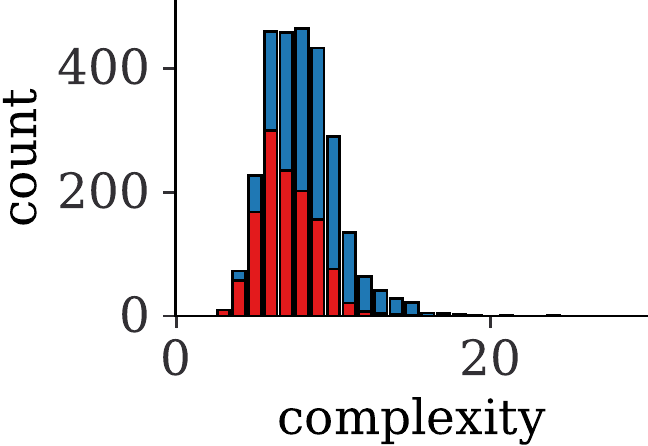}
    \caption{Testset-constant}
\end{subfigure}%
\begin{subfigure}{.25\textwidth}
    \centering
    \includegraphics[width=\linewidth]{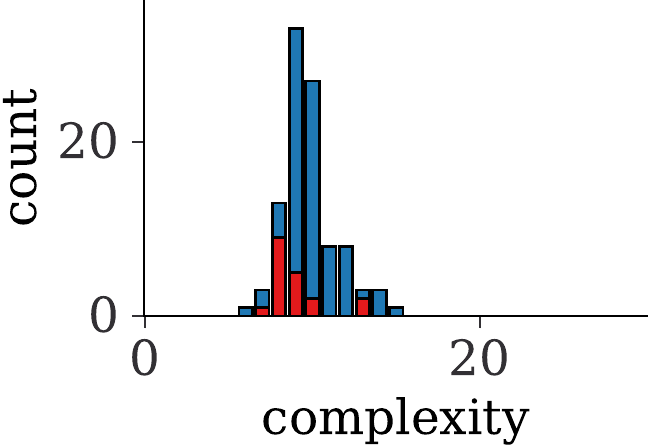}
    \caption{Testset-skeleton}
\end{subfigure}%
\caption{Correctly recovered skeletons by NSODE on testset-iv, testset-constants, and testset-skeletons.}\label{fig:ivconstskelcorrect}
\vspace{-2mm}
\end{figure}

We provide a comprehensive summary of performances of all models on all datasets in \cref{tab:full_result_table}.
Additionally, \cref{fig:classiccorrect,fig:textbookcorrect,fig:163correct,fig:ivconstskelcorrect} show the number of correctly recovered skeletons by each method per complexity.

\begin{table}[h!]
\caption{Detailed performance results for all methods on all (applicable) datasets.}\label{tab:full_result_table}
\centering
\begin{tabularx}{\columnwidth}{l@{\hspace{8pt}}l@{\hspace{8pt}}Y@{\hspace{6pt}}Y@{\hspace{8pt}}Y@{\hspace{0pt}}Y}
\toprule \rowcolor{white}
Dataset                           & Metric                 & NSODE & Sindy & GPLearn & AIFeynman  \\ 
\hline
                                    & skel-recov             & 52.0  & -     &-& -        \\ 
                                  & R$^2 \geq 0.999$             & 28.6  & -     &-& -        \\ 
testset-iv                          & allclose               & 50.6  & -     &-& -        \\ 
                                  & skel-recov \& R$^2 \geq 0.999$  & 17    & -     &-& -        \\ 
                                  & skel-recov \& allclose   & 22.6  & -     &-& -        \\ 
                                  & runtime [s]             & 5.3 & -  & - & - \\
\hline
                                    & skel-recov             & 45.6  & -     &-& -        \\ 
                                  & R$^2 \geq 0.999$              & 21.7  & -     &-& -        \\ 
testset-constant                    & allclose               & 44.7  & -     &-& -        \\ 
                                  & skel-recov \& R$^2 \geq 0.999$  & 9.8   & -     &-& -        \\ 
                                  & skel-recov \& allclose & 16.1  & -     &-& -        \\ 
                                  & runtime [s]             & 5.3 & -  & - & - \\
\hline
                                    & skel-recov             &  19     & -     &-& -        \\ 
                                  & R$^2 \geq 0.999$             &  12     & -     &-& -        \\ 
testset-skel                        & allclose               &  33     & -     &-& -        \\ 
                                  & skel-recov \& R$^2 \geq 0.999$  &  1     & -     &-& -        \\ 
                                  & skel-recov \& allclose &  2    & -     &-& -        \\ 
                                  & runtime [s]             & 5.3 & -  & - & - \\
\hline
                                   & skel-recov             & \bf 37.4  & 3.7   & 2.5 &  14.1   \\ 
                                  & R$^2 \geq 0.999$              & 24.5  &  31.9  & 3.7& \bf 49.7     \\ 
testset-iv-163                      & allclose               & 42.3  & 25.8  & 14.7 & \bf 55.8  \\ 
                                  & skel-recov \& R$^2 \geq 0.999$  & \bf 15.3  & 3.1   & 1.8 &  13.5   \\ 
                                  & skel-recov \& allclose & \bf 15.3  & 3.1   & 1.8  &  13.5  \\ 
                                  & runtime [s]             & 5.4 & \bf0.4 & 29 {\color{gray}+22} & 1203.6 \\
\hline
                                  & skel-recov             &  11.5  &  0  & 3.8  &  \bf 46.2  \\ 
                                  & R$^2 \geq 0.999$              &  57.7  & 57.7  & 23.1 & \bf 88.5   \\ 
classic                             & allclose               &  80.8  & 57.7  & 30.8  & \bf 88.5  \\ 
                                  & skel-recov \& R$^2 \geq 0.999$  & 0     & 0   & 7.7 & \bf46.2 \\ 
                                  & skel-recov \& allclose & 0     & 0   &  7.7  &  \bf46.2  \\ 
                                  & runtime [s]             & 5.2 & \bf0.6 & 23 {\color{gray}+22} & 1291.6 \\
\hline
                                    & skel-recov             & 41.7  & 33.3  & 8.3  & \bf 91.7   \\ 
                                  & R$^2 \geq 0.999$              &  16.7  & 50   & 0.0  & \bf 75   \\ 
textbook                            & allclose               & 25    &  58.3  & 8.3  & \bf 75   \\ 
                                  & skel-recov  \& R$^2 \geq 0.999$ &  33.3   &  41.7  & 0 & \bf 66.7      \\ 
                                  & skel-recov \& allclose & 8.3  &  33.3  & 1.8 & \bf 66.7   \\
                                  & runtime [s]             & 6 & \bf1 & 23 {\color{gray}+22} & 1267.1 \\
\bottomrule
\end{tabularx}%
\end{table}

\end{document}